\documentclass[11pt,letterpaper]{article} 

\usepackage[OT1]{fontenc} 
\usepackage[colorlinks]{hyperref}
\usepackage{url}
\usepackage[numbers]{natbib}

\usepackage{fullpage,times}
\usepackage{amsthm,amsmath,amsfonts} 
\usepackage{epsf,epsfig,caption,subcaption,graphicx} 
\usepackage[ruled,boxed]{algorithm2e}
\usepackage{tikz}
\usepackage{multirow}

\newtheorem{theorem}{Theorem}
\newtheorem{lemma}{Lemma} 
 
\newtheorem{corollary}{Corollary}
\theoremstyle{definition}
\newtheorem{definition}{Definition}

\theoremstyle{remark}

\newtheorem{asss}{Assumption}
\newcommand{\bass}{\begin{asss}}
\newcommand{\eass}{\end{asss}}

\def\exp{\mbox{exp}}

\DeclareMathAlphabet\mathbfcal{OMS}{cmsy}{b}{n}
\begin{document}
\graphicspath{{figs/}}

\begin{center}
	{\bf{\LARGE{Non-parametric sparse additive auto-regressive network models}}}
	
	\vspace*{.1in}
	\begin{tabular}{cc}
		Hao Henry Zhou$^1$\;\;Garvesh Raskutti$^{1,2,3}$\\
	\end{tabular}
	
	\vspace*{.1in}
	
	\begin{tabular}{c}
		$^1$ Department of Statistics, University of Wisconsin-Madison \\
		$^2$ Department of Computer Science\\
		$^3$ Department of Electrical and Computer Engineering \\
	\end{tabular}
	
	\vspace*{.1in}
	
	
\end{center}
\begin{abstract}
 Consider a multi-variate time series $(X_t)_{t=0}^{T}$ where $X_t \in \mathbb{R}^d$ which may represent spike train responses for multiple neurons in a brain, crime event data across multiple regions, and many others. An important challenge associated with these time series models is to estimate an influence network between the $d$ variables, especially when the number of variables $d$ is large meaning we are in the \emph{high-dimensional setting}. Prior work has focused on parametric vector auto-regressive models. However, parametric approaches are somewhat restrictive in practice. In this paper, we use the non-parametric sparse additive model (SpAM) framework to address this challenge. Using a combination of $\beta$ and $\phi$-mixing properties of Markov chains and empirical process techniques for reproducing kernel Hilbert spaces (RKHSs), we provide upper bounds on mean-squared error in terms of the sparsity $s$, logarithm of the dimension $\log d$, number of time points $T$, and the smoothness of the RKHSs. Our rates are sharp up to logarithm factors in many cases. We also provide numerical experiments that support our theoretical results and display potential advantages of using our non-parametric SpAM framework for a Chicago crime dataset. 
\end{abstract}



\section{Introduction}\label{SecIntro}

Multi-variate time series data arise in a number of settings including neuroscience~(\cite{brown2004multiple,DingSchroeder2011}), finance~(\cite{rydberg1999modelling}), social networks~(\cite{chavez2012high,ait2010modeling, zhouZhaSongHawkes}) and others~(\cite{heinen2003modelling,matteson2011forecasting,ogata1999seismicity}). A fundamental question associated with multi-variate time series data is to quantify influence between different players or nodes in the network (e.g. how do firing events in one region of the brain trigger another, how does a change in stock price for one company influence others, e.t.c). To address such a question requires \emph{estimation of an influence network} between the $d$ different players or nodes. Two challenges that arise in estimating such an influence network are (i) developing a suitable network model; and (ii) providing theoretical guarantees for estimating such a network model when the number of nodes $d$ is large.

Prior approaches for addressing these challenges involve parametric approaches ~(\cite{fokianos2011log,fokianos2009poisson,hall16}). In particular,~\cite{hall16} use a generalized linear model framework for estimating the high-dimensional influence network. More concretely, consider samples $(X_t)_{t=0}^T$ where $X_t\in \mathbb R^d$ for every $t$ which could represent continuous data, count data, binary data or others. We define $p(.)$ to be an exponential family probability distribution, which includes, for example,  the Gaussian, Poisson, Bernoulli and others to handle different data types. Specifically, $x\sim p(\theta)$ means that the distribution of the scalar $x$ is associated with the density $p(x|\theta) = h(x)\exp[\varphi(x)\theta-Z(\theta)]$, where $Z(\theta)$ is the so-called log partition function, $\varphi(x)$ is the sufficient statistic of the data, and $h(x)$ is the base measure of the distribution. For the prior parametric approach in~\cite{hall16}, the $j^{th}$ time series observation of $X_{t+1}$ has the following model:
\begin{equation*}\label{eq:par}
	X_{t+1,j}|X_t\sim p(v_j+\sum_{k=1}^{d}A^\ast_{j,k}X_{t,k}),
\end{equation*}
where $A^\ast \in \mathbb{R}^{d \times d}$ is the network parameter of interest. Theoretical guarantees for estimating $A^\ast$ are provided in~\cite{hall16}. One of the limitations of parametric models is that they do not capture non-linear effects such as saturation. Non-parametric approaches are more flexible and apply to broader network model classes but suffer severely from the curse of dimensionality (see e.g.~\cite{Sto85}).

To overcome the curse of dimensionality, the sparse additive models (SpAM) framework was developed (see e.g.~\cite{KolYua10Journal,Meier09,raskutti12,RavLiuLafWas08}). Prior approaches based on the SpAM framework have been applied in the regression setting. In this paper, we consider samples generated from a \emph{non-parametric sparse additive auto-regressive model}, generated by the generalized linear model (GLM), 
\begin{equation}\label{eq:gen}
	X_{t+1,j}|X_t\sim p(v_j+\sum_{k=1}^{d}f^\ast_{j,k}(X_{t,k}))
\end{equation}
where $f^\ast_{j,k}$ is an unknown function belonging to a reproducing kernel Hilbert space $\mathcal H_{j,k}$. The goal is to estimate the $d^2$ functions $(f^\ast_{j,k})_{1 \leq j,k \leq d}$. 

Prior theoretical guarantees for sparse additive models have focused on the setting where samples are independent. In this paper, we analyze the convex penalized sparse and smooth estimator developed and analyzed in~\cite{KolYua10Journal,raskutti12} under the dependent Markov chain model~\eqref{eq:gen}. To provide theoretical guarantees, we assume the Markov chain ``mixes'' using concepts of $\beta$ and $\phi$-mixing of Markov chains. In particular, in contrast to the parametric setting, our convergence rates are a function of $\beta$ or $\phi$ mixing co-efficients, and the smoothness of the RKHS function class. We also support our theoretical guarantees with simulations and show through simulations and a performance analysis on real data the potential advantages of using our non-parametric approach.

\subsection{Our contributions}

As far as we are aware, our paper is the first to provide a theoretical analysis of high-dimensional non-parametric auto-regressive network models. In particular, we make the following contributions.

\begin{itemize}
	\item We provide a scalable non-parametric framework using technologies in sparse additive models for high-dimensional time series models that capture non-linear, non-parametric framework. This provides extensions to prior on high-dimensional parametric models by exploiting RKHSs.
	\item In Section~\ref{SecMain}, we provide the most substantial contribution of this paper which is an upper bound on mean-squared error that applies in the high-dimensional setting. Our rates depend on the sparsity of the function, smoothness of each univariate function, and mixing co-efficients. In particular, our mean-squared error upper bound scales as:
	$$
    \max\big(\frac{s \log d}{\sqrt{mT}}, \sqrt{\frac{m}{T}}\tilde{\epsilon}_m^2 \big),
    $$
	up to logarithm factors, where $s$ is the maximum degree of a given node, $d$ is the number of nodes of the network, $T$ is the number of time points. Here $\tilde{\epsilon}_m$ refers to the univariate rate for estimating a single function in RKHS with $m$ samples (see e.g.~\cite{raskutti12}) and $1 \leq m \leq T$ refers to the number of \emph{blocks} needed depending on the $\beta$ and $\phi$-mixing co-efficients. If the dependence is weak and $m = O(T)$, our mean-squared error bounds are
	optimal up to log factors as compared to prior work on independent models~\cite{raskutti12} while if dependence is strong $m = O(1)$, we obtain the slower rate (up to log factors) of $\frac{1}{\sqrt{T}}$ that is optimal under no dependence assumptions.
	\item We also develop a general proof technique for addressing high-dimensional time series models. Prior proof techniques in ~\cite{hall16} rely heavily on parametric assumptions and constraints on the parameters which allow us to use martingale concentration bounds. This proof technique explicitly exploits
    mixing co-efficients which relies on the well-known ``blocking'' technique for sequences of dependent random variables (see e.g.~\cite{mohri10,nobel93}). In the process of the proof, we also develop upper bounds on Rademacher complexities for RKHSs and other empirical processes under mixing assumptions rather than traditional independence assumptions as discussed in Section~\ref{SecProof}. 
	\item In Section~\ref{SecSimualtions}, we demonstrate through both a simulation study and real data example the flexibility and potential benefit of using the non-parametric approach. In particular we show improved prediction error performance on higher-order polynomials applied to a Chicago crime dataset.
\end{itemize}

The remainder of the paper is organized as follows. In Section~\ref{SecNotPrelim}, we introduce
the preliminaries for RKHSs, and beta-mixing of Markov chains. In Section~\ref{SecModel}, we present the non-parametric multi-variate auto-regressive network model and its estimating scheme. In Section~\ref{SecMain}, we present the main
theoretical results and focus on specific cases of finite-rank kernels and Sobolev
spaces. In Section~\ref{SecProof}, we provide the main steps of the proof, deferring
the more technical steps to the appendix and in Section~\ref{SecSimualtions}, we provide
a simulation study that supports our theoretical guarantees and a performance analysis on Chicago crime data.

\section{Preliminaries}

\label{SecNotPrelim}

In this section,  we introduce the basic concepts of RKHSs, and then the standard definitions of $\beta$ and $\phi$ mixing for stationary processes.

\subsection{Reproducing Kernel Hilbert Spaces}

First we introduce the basics for RKHSs. Given a subset $\mathcal X \subset \mathbb R$ and a probability measure $\mathbb Q$ on $\mathcal X$, we consider a Hilbert space $\mathcal H \subset \mathcal L^2(\mathbb Q)$, meaning a family of functions $g : \mathcal X \rightarrow \mathbb R$, with $\|g\|_{\mathcal L^2(\mathbb Q)} < \infty$, and an associated inner product $\langle \cdot,\cdot \rangle_{\mathcal H}$ under which $\mathcal H$ is complete. The space $\mathcal H$ is a reproducing kernel Hilbert space (RKHS) if there exists a symmetric function $\mathcal K : \mathcal X \times \mathcal X \rightarrow \mathbb R_+$ such that: (a) for each $x \in \mathcal X$ , the function $\mathcal K(x,\cdot)$ belongs to the Hilbert space $\mathcal H$, and (b) we have the reproducing relation $g(x) = \langle g, \mathcal K(x,\cdot)\rangle_{\mathcal H}$ for all $g \in \mathcal H$. Any such kernel function must be positive semidefinite; under suitable regularity conditions, Mercer’s theorem \cite{mercer1909} guarantees that the kernel has an eigen-expansion of the form
$$\mathcal K(x,x')=\sum_{i=1}^{\infty}\mu_i\Phi_i(x)\Phi_i(x')$$
where $\mu_1\geq \mu_2\geq \mu_3\geq ... \geq 0$ are a non-negative sequence of eigenvalues, and $\{\Phi_i\}_{i=1}^{\infty}$ are the associated eigenfunctions, taken to be orthonormal in $\mathcal L^2(\mathbb Q)$. The decay rate of these eigenvalues will play a crucial role in our analysis, since they ultimately determine the rate $\epsilon_m,\tilde{\epsilon}_m$ (to be specified later) for the univariate RKHS’s in our function classes.

Since the eigenfunctions $\{\Phi_i\}_{i=1}^{\infty}$ form an orthonormal basis, any function $g\in \mathcal H$ has an expansion of the $g(x)=\sum_{i=1}^{\infty}a_i\Phi_i(x)$, where $a_i=\langle g,\Phi_i\rangle_{\mathcal L^2(\mathbb Q)}=\int_{\mathcal X} g(x)\Phi_i(x)d \mathbb Q(x)$ are (generalized) Fourier coefficients. Associated with any two functions in $\mathcal H$, say $g(x)=\sum_{i=1}^{\infty}a_i\Phi_i(x)$ and $f(x)=\sum_{i=1}^{\infty}b_i\Phi_i(x)$ are two distinct inner products. The first is the usual inner product in the space $\mathcal L^2(\mathbb Q)$-namely, $\langle g,f\rangle_{\mathcal L^2(\mathbb Q)}:=\int_{\mathcal X}g(x)f(x) d \mathbb Q(x)$. By Parseval's theorem, it has an equivalent representation in terms of the expansion coefficients, namely
$$\langle g,f\rangle_{\mathcal L^2(\mathbb Q)}=\sum_{i=1}^{\infty}a_ib_i.$$
The second inner product, denoted $\langle g,f\rangle_{\mathcal H}$ is the one that defines the Hilbert space which can be written in terms of the kernel eigenvalues and generalized Fourier coefficients as
$$\langle g,f\rangle_{\mathcal H}=\sum_{i=1}^{\infty}\frac{a_ib_i}{\mu_i}.$$
For more background on reproducing kernel Hilbert spaces, we refer the reader to various standard references \cite{aronszajn1950,saitoh1988,smola1998,wahba1990,weinert1982}.

Furthermore, for the subset $S_j \in \{1,2,..,d\}$, let $f_j := \sum_{k \in S_j}{f_{j, k} (x_k)}$, where $x_k \in \mathcal X$ and $f_{j, k} \in \mathcal{H}_{j,k}$ is the RKHS that $f_{j, k}$ lies in. Hence we define the norm 
$$\|f_j\|^2_{\mathcal{H}_j(S_j)}:=\sum_{k\in S_j}\|f_{j,k}\|_{\mathcal{H}_{j,k}}^2,$$
where $\|\cdot\|_{\mathcal{H}_{j,k}}$ denotes the norm on the univariate Hilbert space $\mathcal{H}_{j,k}$.

\subsection{Mixing}
Now we introduce standard definitions for dependent observations based on mixing theory \cite{doukhan1994} for stationary processes.

\begin{definition}
	A sequence of random variables $Z=\{Z_t\}_{t=0}^{\infty}$ is said to be \emph{stationary} if for any $t_0$ and non-negative integers $t_1$ and $t_2$, the random vectors $(Z_{t_0},...,Z_{t_0+t_1})$ and $(Z_{t_0+t_2},...,Z_{t_0+t_1+t_2})$ have the same distribution.
\end{definition}

Thus the index $t$ or time, does not affect the distribution of a variable $Z_t$ in a stationary sequence. This does not imply independence however and we capture the dependence through mixing conditions. 
The following is a standard definition giving a measure of the dependence of the random variables $Z_t$ within a stationary sequence. There are several equivalent definitions of these quantities, we are adopting here a version convenient for our analysis, as in~\cite{mohri10,yu1994}.

\begin{definition}
	Let $Z=\{Z_t\}_{t=0}^{\infty}$ be a stationary sequence of random variables. For any $i_1,i_2\in Z\cup\{0,\infty\}$, let $\sigma_{i_1}^{i_2}$ denote the $\sigma$-algebra generated by the random variables $Z_t,i_1\leq t\leq i_2$. Then, for any positive integer $\ell$, the $\beta$-mixing and $\phi$-mixing coefficients of the stochastic process $Z$ are defined as
	\begin{equation*}
	\beta(\ell)=\sup_{t}E_{B\in\sigma_0^t}[\sup_{A\in\sigma_{t+\ell}^\infty}|P[A|B]-P[A]|]\ ,\phi(\ell)=\sup_{t,A\in \sigma_{t+\ell}^\infty,B\in\sigma_0^t}|P[A|B]-P[A]|.
	\end{equation*}
	$Z$ is said to be $\beta$-mixing ($\phi$-mixing) if $\beta(\ell)\rightarrow 0$ (resp. $\phi(\ell)\rightarrow 0$) as $\ell\rightarrow \infty$. Furthermore $Z$ is said to be \emph{algebraically} $\beta$-mixing (algebraically $\phi$-mixing) if there exist real numbers $\beta_0>0$ (resp. $\phi_0>0$) and $r>0$ such that $\beta(\ell)\leq \beta_0/\ell^r$ (resp. $\phi(\ell)\leq \phi_0/\ell^r$) for all $\ell$.
\end{definition}

Both $\beta(\ell)$ and $\phi(\ell)$ measure the dependence of an event on those that occurred more than $\ell$ units of time in the past. $\beta$-mixing is a weaker assumption than $\phi$-mixing and thus includes more general non-i.i.d. processes.\\
\section{Model and estimator} \label{SecModel}
In this section, we introduce the sparse additive auto-regressive network model and the sparse and smooth regularized schemes that we implement and analyze.

\subsection{Sparse additive auto-regressive network model}


From Equation~\eqref{eq:gen} in Section~\ref{SecIntro}, we can state the conditional distribution explicitly as:
\begin{equation}
\mathbb P(X_{t+1}|X_t)  = \prod_{j=1}^d h(X_{t+1,j})\exp\left\{\varphi(X_{t+1,j})(v_j+\sum_{k=1}^d f^\ast_{j,k}(X_{t,k}))-Z(v_j+\sum_{k=1}^d f^\ast_{j,k}(X_{t,k}))\right\}
\end{equation}
where $f^\ast_{j,k}$ is an unknown function belonging to a RKHS $\mathcal H_{j,k}$, $v\in [v_{min},v_{max}]^{d}$ are known constant offset parameters. Recall that $Z(\cdot)$ refers to the log-partition function and $\varphi(\cdot)$ refers to the sufficient statistic. This model has the Markov and conditional independence properties, that is, conditioning on the previous data at time point $t-1$, the elements of $X_t$ are independent of one another and $X_t$ are independent with data before time $t-1$. We note that while we assume that $v$ is a known constant vector, if we assume there is some unknown constant offset that we would like to estimate, we can fold it into the estimation of $f^\ast$ via appending a constant 1 column in $X_t$. 

We assume that the data we observe is $(X_t)_{t=0}^T$ and our goal is to estimate $f^\ast$, which is constructed element-wise by $f^\ast_{j,k}$. However, in our setting where $d$ may be large, the sample size $T$ may not be sufficient even under the additivity assumption and we need further structural assumptions. Hence we assume that the network function $f^\ast$ is sparse which does not have too many non-zero functions.
To be precise, we define the sparse supports $(S_1, S_2,...,S_d)$ as:
$$S_j\subset \{1,2,...,d\}\text{, for any }j=1,2,...,d.$$
We consider network function $f^\ast$ is only non-zero on supports $\{S_j\}_{j=1}^d$, which means
\begin{equation*}\begin{aligned}
f^\ast\in \mathcal{H}(S):=\{f_{j,k}\in \mathcal{H}_{j,k}|f_{j,k}=0\mbox{ for any }k\notin S_j\}.
\end{aligned}\end{equation*}
The support $S_j$ is the set of nodes that influence node $j$ and $s_j = |S_j|$ refers to the \emph{in-degree} of node $j$. In this paper we assume that the function matrix $f^\ast$ is $s$-sparse, meaning that $f^\ast$ belongs to $\mathcal{H}(S)$ where $|S|=\sum_{j=1}^d |S_j|\leq s$. From a network perspective, $s$ represents the total number of edges in the network.

\subsection{Sparse and smooth estimator}

The estimator that we analyze in this paper is the standard sparse and smooth estimator developed in~\cite{KolYua10Journal,raskutti12}, for each node $j$. To simplify notation and without loss of generality, in later statements we assume $\mathcal H_{j,k}$ refers to the same RKHS $\mathcal H$, and define $\mathcal H_j=\{f_j|f_j=\sum_{k=1}^d f_{j,k},\text{ for any }f_{j,k}\in \mathcal H\}$ which corresponds to the additive function class for each node $j$. Further we define the \emph{empirical norm} $\|f_{j,k}\|^2_{T}:=\frac{1}{T}\sum_{t=0}^T f_{j,k}^2(X_{t,k})$. For any function of the form $f_j=\sum_{k=1}^d f_{j,k}$, the $(L^2(\mathbb P_T),1)$ and $(\mathcal H,1)$-norms are given by
$$\|f_j\|_{T,1}=\sum_{k=1}^d \|f_{j,k}\|_{T},\ \mbox{and}\  \|f_j\|_{\mathcal H,1}=\sum_{k=1}^d \|f_{j,k}\|_{\mathcal H}$$
respectively. Using this notation, we estimate $f^\ast_j$ via a regularized maximum likelihood estimator (RMLE) by solving the following optimization problem, for any $j\in \{1,2,..,d\}$:
\begin{equation}\label{eq:optimfun}
\hat{f}_j =\arg\min_{f_j \in \mathcal H_j}\frac{1}{2T}\sum_{t=0}^{T} \left(Z(v_j+f_j(X_t))-(v_j+f_j(X_t))\varphi(X_{t+1,j})\right)+\lambda_T\| f_j\|_{T,1} +\lambda_H \| f_j\|_{\mathcal H,1}.
\end{equation}

Here $(\lambda_T,\lambda_H)$ is a pair of positive regularization parameters whose choice will be specified by our theory. An attractive feature of this optimization problem is that, as a straightforward consequence of the representer theorem \cite{kimeldorf1971,smola1998}, it can be reduced to an equivalent convex program in $\mathbb R^T\times \mathbb R^{d^2}$. In particular, for each $(j,k)\in \{1,2,...,d\}^2$, let $\mathcal K$ denote the kernel function associated with RKHS $\mathcal H$ where $f_{j,k}$ belongs to. We define the collection of empirical kernel matrices $\mathbb K^{j,k}\in \mathbb R^{T\times T}$ with entries $\mathbb K^{j,k}_{t_1,t_2}=\mathcal K(X_{t_1,k},X_{t_2,k})$. As discussed in~\cite{KolYua10Journal,raskutti12}, by the representer theorem, any solution $\hat{f}_j$ to the variational problem can be expressed in terms of a linear expansion of the kernel matrices,
$$\hat{f}_j(z)=\sum_{k=1}^d \sum_{t=1}^T \hat{\alpha}_{j,k,t}\mathcal K(z_k,X_{t,k})$$
for a collection of weights $\{\hat{\alpha}_{j,k}\in \mathbb R^T\;\;(j,k)\in\{1,2,..,d\}^2\}$. The optimal weights are obtained by solving the convex problem
$$\hat{\alpha}_j=(\hat{\alpha}_{j,1},...,\hat{\alpha}_{j,d})=arg\min_{\alpha_{j,k}\in \mathbb R^T}\frac{1}{2T}\sum_{t=0}^{T} \left(Z(v_j+\sum_{k=1}^d\mathbb K^{j,k}\alpha_{j,k})-(v_j+\sum_{k=1}^d\mathbb K^{j,k}\alpha_{j,k})\varphi(X_{t+1,j})\right)$$
$$+\lambda_T\sum_{k=1}^d\sqrt{\frac{1}{T}\|\mathbb K^{j,k}\alpha_{j,k}\|_2^2} +\lambda_H \sum_{k=1}^d\sqrt{\alpha_{j,k}^T \mathbb K^{j,k}\alpha_{j,k}}.$$
This problem is a second-order cone program (SOCP), and there are various algorithms for solving it to arbitrary accuracy in polynomial time of $(T,d)$, among them interior point methods (e.g., see the book \cite{boyd2004}). 

Other more computationally tractable approaches for estimating sparse additive models have been developed in~\cite{Meier09,RavLiuLafWas08} and in our experiments section we use the package ``SAM'' based on the algorithm developed in~\cite{RavLiuLafWas08}. However from a theoretical perspective the sparse and smooth SOCP defined above has benefits since it is the only estimator with provably minimax optimal rates in the case of independent design (see e.g.~\cite{raskutti12}).\\

\section{Main results}\label{SecMain}
In this section, we provide the main general theoretical results. In particular, we derive error bounds on $\|\hat{f}-f^\ast\|_T^2$, the difference in empirical $\mathcal L_2(\mathbb P_T)$ norm between the regularized maximum likelihood estimator, $\hat{f}$, and the true generating network, $f^\ast$, under the assumption that the true network is $s$-sparse. 

First we incorporate the smoothness of functions in each RKHS $\mathcal{H}$. We refer to $\epsilon_m$ as the critical univariate rate, which depends on the Rademacher complexity of each function class. That $\epsilon_m$ is defined as the minimal value of  $\sigma$, such that
\begin{equation*}
  \frac{1}{\sqrt{m}}\sqrt{\sum_{i=1}^{\infty}\min(\mu_i,\sigma^2)}\leq \sigma^2,
\end{equation*}
where $\{\mu_i\}_{i=1}^\infty$ are the eigenvalues in Mercer's decomposition of the kernel related to the univariate RKHS (see~\cite{mercer1909}). In our work, we define $\tilde{\epsilon}_m$ as the univariate rate for a slightly modified Rademacher complexity, which is the minimal value of $\sigma$, such that there exists a $M_0\geq 1$ satisfying
\begin{equation*}
	 \log(dT)\{3\frac{\log(M_0dT)}{\sqrt{m}}\sqrt{\sum_{i=1}^{M_0}\min(\mu_i,\sigma^2)}+\sqrt{\frac{T}{m}}\sqrt{\sum_{M_0+1}^{\infty}\min(\mu_i,\sigma^2)}\}\leq \sigma^2.
\end{equation*}
{\it Remark.} Note that since the left side of the inequality for $\tilde{\epsilon}_m$ is always larger than it for $\epsilon_m$, the definitions of $\tilde{\epsilon}_m$ and $\epsilon_m$ tell us that $\epsilon_m\leq\tilde{\epsilon}_m$.
Furthermore $\tilde{\epsilon}_m$ is of order $O(\epsilon_{m}*\log(dT)^2)$ for finite rank kernel and kernel with decay rate $i^{-2\alpha}$. See Subsection~\ref{section:twoexamples} for more details. The modified definition $\tilde{\epsilon}_m$ allows us to extend the error bounds on $\|\hat{f}-f^\ast\|_T^2$ to the dependent case at the price of additional log factors.

\subsection{Assumptions}

 We first state the assumptions in this subsection and then present our main results in the next subsection. Without loss of generality (by re-centering the functions as needed), we assume that
$$\mathbb E[f_{j,k}(X_{t,j})]=\int_{\mathcal X} f_{j,k}(x) d \mathbb P(x)=0 \text{ for all } f_{j,k} \in \mathcal H_{j,k}\text{, all }t.$$ 
Besides, for each $(j,k)\in \{1,...,d\}^2$, we make the minor technical assumptions:
\begin{itemize}
	\item For any $f_{j,k}\in \mathcal H$, $\| f_{j,k}\|_{\mathcal H}\leq 1$ and $\| f_{j,k}\|_{\infty}\leq 1$.
	\item For any $\mathcal H$, the associated eigenfunctions in Mercer’s decomposition $\{\Phi_i\}_{i=1}^{\infty}$ satisfy $\sup_x|\Phi_i(x)|\leq 1$ for each $i=1,...,\infty$.
\end{itemize}
The first condition is mild and also assumed in~\cite{raskutti12}. The second condition is satisfied by the bounded basis, for example, the Fourier basis. We proceed to the main assumptions by denoting $s_{\max} = \max_j s_j$ as the maximum in-degree of the network and denoting $\mathcal{H}_{\mu} = \sum_{i=1}^{\infty}{\mu_i}$ as the trace of the RKHS $\mathcal H$.

\bass [Bounded Noise]\label{asp:boundednoise}
 Let $w_{t,j}=\frac{1}{2}(\varphi(X_{t+1,j})- Z'(v_j+f^\ast_{j}(X_t)))$, we assume that $E[w_{t,j}]=0$ and with high probability $w_{t,j}\in [-\log(dT),\log(dT)]$, for any  $j\in \{1,2,...,d\},t=1,2,...,T$. 
\eass

{\it Remark.} It can be checked that for (1) Gaussian link function with bounded noise or (2) Bernoulli link function, $w_{t,j}=O(1)$ with probability $1$. For other generalized linear model cases, such as (1) Gaussian link function with Gaussian noise or (2) Poisson link function under the assumption $f^\ast_{j,k}\leq 0$ for any $(j,k)$, we have that $|w_{t,j}|\leq C\log(dT)$ with probability at least $1-\exp(-c\log(dT))$ for some constants $C$ and $c$ 
(see the proof of Lemma 1 in \cite{hall16}). 

\bass[Strong Convexity]\label{asp:strongconv}
For any $x,y$ in an interval $(v_{\min}-a,v_{\max}+a)$, $$\vartheta\|x-y\|^2\leq [Z(x)-Z(y)-Z'(y)(x-y)]$$.
\eass

{\it Remark.} For the Gaussian link function, $a = \infty$ and $\vartheta=1$. For Bernoulli link function, $a = (16\sqrt{\mathcal{H}_{\mu}}+1)s_{\max}$ and  $\vartheta=(e^{(\max(v_{\max},-v_{\min})+(16\sqrt{\mathcal{H}_{\mu}}+1)s_{\max})}+3)^{-1}$. For Poisson link function, $a = (16\sqrt{\mathcal{H}_{\mu}}+1)s_{\max}$ and $\vartheta=e^{v_{\min}-(16\sqrt{\mathcal{H}_{\mu}}+1)s_{\max}}$ where recall that $s_{\max}$ is the maximum in-degree of the network.

 
\bass[Mixing]\label{asp:mix}
The sequence $(X_t)_{t=0}^{\infty}$ defined by the model~\eqref{eq:gen} is a stationary sequence satisfying one of the following mixing conditions:
\begin{itemize}
\item[(a)] $\beta$-mixing with $r_\beta > 1$.
\item[(b)] $\phi$-mixing with $r_\phi\geq 0.781$.
\end{itemize}
\eass
We can show a tighter bound when $r_\phi\leq 2$ using the concentration inequality from~\cite{kontorovich08}. The condition $r_\phi\geq 0.781$ arises from the technical condition in which $(r_\phi+2)\times(2r_\phi-1)\geq 2r_\phi$ (see the Proof of Lemma~\ref{lemma:newunivariate}). Numerous results in the statistical machine learning literature rely on knowledge of the $\beta$-mixing coefficient~\cite{mcdonald11,vidyasagar02}. Many common time series models are known to be $\beta$-mixing, and the rates of decay are known given the true parameters of the process, for example, ARMA models, GARCH models, and certain Markov processes~\cite{mokkadem88, carrasco02, doukhan91}. The $\phi$-mixing condition is stronger but as we observe later allows a sharper mean-squared error bound. 

\bass[Fourth Moment Assumption]\label{asp:fourth}
$E[g^4(x)]\leq CE[g^2(x)]$ for some constant $C$, for all $g\in \mathcal F_j:= \cup_{|S_j|=s_j}H_j(S_j)$, for any $j\in\{1,2,..,d\}$ where the expectation is taken over $\mathbb{Q}$.
\eass

Note that Assumption~\ref{asp:fourth} is a technical assumption also required in~\cite{raskutti12} and is satisfied under mild dependence across the covariates.

\subsection{Main Theorem}\label{sec:3}

Before we state the main result, we discuss the choice of tuning parameters $\lambda_T$ and $\lambda_H$.

{\bf Optimal tuning parameters:} Define $\gamma_m=c_1\max\left(\epsilon_m,\sqrt{\frac{\log (dT)}{m}}\right)$, where $c_1>0$ is a sufficiently large constant, independent of $T$, $s$ and $d$, and $m\gamma_m^2=\Omega(-\log(\gamma_m))$ and $m\gamma_m^2\rightarrow \infty$ as $m\rightarrow \infty$. $\tilde\gamma_m=\max(\gamma_m,\tilde{\epsilon}_m)$. The parameter $m$ is a function of $T$ and is defined in Thm.~\ref{thm:key} and Thm.~\ref{thm:follow}. Then we have the following optimal choices of tuning parameters:
$$\lambda_T\geq 8\sqrt{2}\sqrt{\frac{m}{T}}\tilde\gamma_m, \lambda_H\geq 8\sqrt{2}\sqrt{\frac{m}{T}}\tilde\gamma_m^2,$$
$$\lambda_T=O(\sqrt{\frac{m}{T}}\tilde\gamma_m), \lambda_H=O(\sqrt{\frac{m}{T}}\tilde\gamma_m^2).$$
Clearly it is possible to choose larger values of $\lambda_T$ and $\lambda_H$ at the expense of slower rates.

\begin{theorem}\label{thm:key}
	Under Assumptions~\ref{asp:boundednoise}, ~\ref{asp:strongconv}, ~\ref{asp:mix} (a), and \ref{asp:fourth}. Then there exists a constant $C$ such that for each $1 \leq j \leq d$,
	\begin{equation}
	\|\hat{f}_j-f_j^{\ast}\|_T^2\leq C\frac{s_j}{\vartheta^2}\left(\frac{\log (dT)}{\sqrt{mT}}+\sqrt{\frac{m}{T}}\tilde{\epsilon}_m^2\right),
	\end{equation}
with probability at least $1-\frac{1}{T}-\left(c_2\exp(-c_3m\gamma_m^2)+T^{-\left(\frac{1-c_0}{c_0}\right)}\right)$, where $m=T^{\frac{c_0r_\beta-1}{c_0r_\beta}}$ for $\beta$-mixing when $r_\beta\geq 1/c_0$, and $c_2$ and $c_3$ are constants. The parameter $c_0$ can be any number between $0$ and $1$.
\end{theorem}

\begin{itemize}
\item Note that the term $\tilde{\epsilon}_m^2$ accounts for the smoothness of the function class, $\vartheta$ accounts for the smoothness of the GLM loss, and $m$ denotes the degree of dependence in terms of the number of blocks in $T$ samples.
\item In the very weakly dependent case $r_{\beta} \rightarrow \infty$ and $m = O(T)$, and we recover the standard rates for sparse additive models $\frac{s_j \log d}{T} + s_j \tilde{\epsilon}_T^2$ (see e.g.~\cite{raskutti12}) up to logarithm factors. In the highly dependent case $m = O(1)$, we end up with a rate proportional to $\frac{1}{\sqrt{T}}$ (up to log factors in terms of $T$ only) which is consistent with the rates for the lasso under no independence assumptions.
\item Note that we have provided rates on the difference of functions $\hat{f}_j-f_j^{\ast}$ for each $1 \leq j \leq d$. To obtain rates for the whole network function $\hat{f}-f^{\ast}$, we simply add up the errors and note that $s = \sum_{j=1}^d{s_j}$.
\item To compare to upper bounds in the parametric case in ~\cite{hall16}, if $m = O(T)$ and $\tilde{\epsilon}_m^2 = O(\frac{1}{m})$, we obtain the same rates. Note however that in ~\cite{hall16} we require strict assumptions on the network parameter instead of the mixing conditions we impose here.
\item A larger $c_0$ leads to a larger $m$ and a lower probability from the term $T^{-\frac{1-c_0}{c_0}}$.
\end{itemize}

When $r_\phi\geq 2$, Theorem~\ref{thm:key} on $\beta$-mixing directly implies the results for $\phi$-mixing. When $0.781\leq r_\phi\leq 2$, we can present a tighter result using the concentration inequality from \cite{kontorovich08}.
\begin{theorem}\label{thm:follow}
	Under same assumptions as in Thm.~\ref{thm:key}, if we assume $\phi$-mixing when $0.781\leq r_\phi\leq 2$, then there exists a constant $C$ such that for each $1 \leq j \leq d$,
	\begin{equation}
	\|\hat{f}_j-f_j^{\ast}\|_T^2\leq C\frac{s_j}{\vartheta^2}\left(\frac{\log (dT)}{\sqrt{mT}}+\sqrt{\frac{m}{T}}\tilde{\epsilon}_m^2\right),
	\end{equation}
	with probability at least $1-\frac{1}{T}-c_2\exp(-c_3(m\gamma_m^2)^2)$, where $m=T^{\frac{r_\phi}{r_\phi+2}}$ for $\phi$-mixing when $0.781\leq r_\phi\leq 2$, $c_2$ and $c_3$ are constants.  
\end{theorem}

Note that $m = T^{\frac{r_\phi}{r_\phi+2}}$ is strictly larger than $m = T^{\frac{r_\phi-1}{r_\phi}}$ for $r_\phi\leq 2$ which is why Theorem~\ref{thm:follow} is a sharper result.

\subsection{Examples}\label{section:twoexamples}
We now focus on two specific classes of functions, finite-rank kernels and infinite-rank kernels with polynomial decaying eigenvalues. First, we discuss finite ($\xi$) rank operators, meaning that the kernel function can be expanded in terms of $\xi$ eigenfunctions. This class includes linear functions, polynomial functions, as well as any function class where functions have finite basis expansions. 

\begin{lemma}\label{example:1}
For a univariate kernel with finite rank $\xi$, $\tilde{\epsilon}_m=O\left(\sqrt{\frac{\xi}{m}}\log^2(\xi dT)\right)$.
\end{lemma}

Using Lemma~\ref{example:1} and $\epsilon_m$ calculated from \cite{raskutti12} gives us the following result. Note that for $T = O(m)$, we end up with the usual parametric rate.

\begin{corollary}\label{Cor:1}
Under the same conditions as Theorem \ref{thm:key}, consider a univariate kernel with finite rank $\xi$. Then there exists a constant $C$ such that for each $1\leq j\leq d$,
\begin{equation}
\|\hat{f}_j-f_j^{\ast}\|_T^2\leq C\frac{s_j}{\vartheta^2}\frac{\xi}{\sqrt{mT}}\log^4(\xi dT),
\end{equation}
with probability at least $1-\frac{1}{T}-\left(c_2\exp(-c_3(\xi+\log d))+T^{-\left(\frac{1-c_0}{c_0}\right)}\right)$, where
$m=T^{\frac{c_0r_\beta-1}{c_0r_\beta}}$ for $\beta$-mixing when $r_\beta\geq 1/c_0$, $c_2$ and $c_3$ are constants. 
\end{corollary}

Next, we present a result for the RKHS with infinitely many eigenvalues, but whose eigenvalues decay at a rate $\mu_\ell=(1/\ell)^{2\alpha}$ for some parameter $\alpha\geq 1/2$. Among other examples, this includes Sobolev spaces, say consisting of functions with $\alpha$ derivatives (e.g., \cite{birman67,gu13}).

\begin{lemma}\label{example:2}
	For a univariate kernel with eigenvalue decay $\mu_\ell=(1/\ell)^{2\alpha}$ for some $\alpha\geq 1/2$, we have that 
	$\tilde{\epsilon}_m= O\left(\left(\frac{\log^2(dT)}{\sqrt{m}}\right)^{\frac{2\alpha}{2\alpha+1}}\right)$.
\end{lemma}

\begin{corollary}
Under the same conditions as Theorem \ref{thm:key}, consider a univariate kernel with eigenvalue decay $\mu_\ell=(1/\ell)^{2\alpha}$ for some $\alpha\geq 1/2$. Then there exists a constant $C$ such that for each $1\leq j\leq d$,
\begin{equation}
\|\hat{f}_j-f_j^{\ast}\|_T^2\leq C\frac{s_j}{\vartheta^2}\frac{\log^{\frac{8\alpha}{2\alpha+1}}(dT)}{\sqrt{m^{\frac{2\alpha-1}{2\alpha+1}}T}},
\end{equation}
with probability at least $1-\frac{1}{T}-T^{-\left(\frac{1-c_0}{c_0}\right)}$, where $m=T^{\frac{c_0r_\beta-1}{c_0r_\beta}}$ for $\beta$-mixing when $r_\beta\geq 1/c_0$. 
\end{corollary}

Note that if $m = O(T)$, we obtain the rate $O(\frac{s_j}{T^{\frac{2 \alpha}{2 \alpha+1}}})$ up to log factors which is optimal in the independent case.
\section{Proof for the main result (Theorem~\ref{thm:key})}\label{SecProof}

At a high level, the proof for Theorem~\ref{thm:key}) follows similar steps to the proof of Theorem 1 in ~\cite{raskutti12}. However a number of additional challenges arise when dealing with dependent data. The key challenge in the proof is that the traditional results for Rademacher complexities of RKHSs and empirical processes assume independence and do not hold for dependent processes. These problems are addressed by Theorem~\ref{theorem:Rademacher} and Theorem~\ref{thm:univariate} in this work.


\subsection{Establishing the basic inequality}
Our goal is to estimate the accuracy of $f_j^\ast(\cdot)$ for every integer $j$ with $1 \leq j \leq d$. We denote the expected $\mathcal L_2(\mathbb P)$ norm of a function $g$ as $\|g\|^2_{2}=\mathbb E\| g\|^2_T$ where the expectation is taken over the distribution of $(X_t)_{t=0}^{T}$. We begin the proof by establishing a basic inequality on the error function $\Delta_j(.)=\hat{f}_j(.)-f_j^{\ast}(.)$. Since $\hat{f}_j(.)$ and $f_j^\ast$ are, respectively, optimal and feasible for \eqref{eq:optimfun}, we are guaranteed that
\begin{equation*}\begin{aligned}
& \frac{1}{2T}\sum_{t=1}^{T} (Z(v_j+\hat{f}_j(X_t))-(v_j+\hat{f}_j(X_t))\varphi(X_{t+1,j}))+\lambda_T\| \hat{f}_j\|_{T,1} +\lambda_H \| \hat{f}_j\|_{H,1}\\
& \leq \frac{1}{2T}\sum_{t=1}^{T} (Z(v_j+f^\ast_j(X_t))-(v_j+f^\ast_j(X_t))\varphi(X_{t+1,j}))+\lambda_T\| f^\ast_j\|_{T,1} +\lambda_H \| f^\ast_j\|_{H,1}.
\end{aligned}\end{equation*}
Using our definition $w_{t,j} = \frac{1}{2}(\varphi(X_{t+1,j})-E[\varphi(X_{t+1,j})|X_t]) = \frac{1}{2}(\varphi(X_{t+1,j})-Z'(v_j+f^\ast_j(X_t)))$, that is
\begin{equation*}\begin{aligned}
& \frac{1}{2T}\sum_{t=1}^{T} (Z(v_j+\hat{f}_j(X_t))-\hat{f}_j(X_t)(Z'(v_j+f^\ast_j(X_t))+2w_{t,j}))+\lambda_T\| \hat{f}_j\|_{T,1} +\lambda_H \| \hat{f}_j\|_{H,1}\\
& \leq \frac{1}{2T}\sum_{t=1}^{T} (Z(v_j+f^\ast_j(X_t))-f^\ast_j(X_t)(Z'(v_j+f^\ast_j(X_t))+2w_{t,j}))+\lambda_T\| f^\ast_j\|_{T,1} +\lambda_H \| f^\ast_j\|_{H,1}.
\end{aligned}\end{equation*}
Let $B_Z(\cdot||\cdot)$ denote the Bregman divergence induced by the strictly convex function $Z$, some simple algebra yields that
\begin{equation}\label{eq:basicineq}
\frac{1}{2T}\sum_{t=1}^{T} B_Z(v_j+\hat{f}_j(X_t)||v_j+f^\ast_j(X_t))\leq \frac{1}{T}\sum_{t=1}^{T} \Delta_j(X_t)w_{t,j}+\lambda_T\| \Delta_j\|_{T,1} +\lambda_H \| \Delta_j\|_{H,1}
\end{equation}
which we refer to as our basic inequality (see e.g.~\cite{geer2000} for more details on the basic inequality).
\subsection{Controlling the noise term}
Let $\Delta_{j,k}(\cdot)=\hat{f}_{j,k}(\cdot)-f^\ast_{j,k}(\cdot)$ for any $k=1,2,...,d$. Next, we provide control for the right-hand side of inequality \eqref{eq:basicineq} by bounding the Rademacher complexity for the univariate functions in terms of their $L^2(\mathbb P_T)$ and $\mathcal H$ norms. We point out that tools required for such control are not well-established in the dependent case which means that we first establish the Rademacher complexity result (Theorem~\ref{theorem:Rademacher}) and the uniform convergence rate for averages in the empirical process (Theorem~\ref{thm:univariate}) for the dependent case (results for the independent case are provided as Lemma 7 in~\cite{raskutti12}).
\begin{theorem}[Rademacher complexity]\label{theorem:Rademacher}
	Under Assumption~\ref{asp:boundednoise}, define the event
	\begin{equation*}
	\mathcal A_{m,T}=\left\{\forall (j,k)\in \{1,2,...,d\}^2,\forall \sigma\geq \tilde{\epsilon}_m,\sup_{\| f_{j,k} \|_{\mathcal H}\leq 1, \| f_{j,k}\|_2\leq \sigma} \left| \frac{1}{T}\sum_{t=1}^T f_{j,k}(X_t)w_{t,j} \right|\leq \sqrt{2}\sqrt{\frac{m}{T}}\sigma^2\right\}.
	\end{equation*}
	Then $\mathbb P(\mathcal A_{m,T})\geq1-\frac{1}{T}$.
\end{theorem}

{\it Remark.} We have a correction term $\sqrt{\frac{T}{m}}$ for $m<T$, in order to connect our Rademacher complexity result with mixing conditions. In the independnet case, $m=T$ which has been proven in prior work. \\


\begin{theorem}\label{thm:univariate}
	Define the event
	\begin{equation}\label{eq:lemmaunivariate}
		\mathcal B_{m,T}=\left\{\sup_{j,k}\sup_{f_{j,k}\in B_H(1), \|f_{j,k}\|_2\leq \gamma_m} |\|f_{j,k}\|_T-\|f_{j,k}\|_2|\leq \frac{\gamma_m}{2}\right\}
	\end{equation}
where $m=T^{\frac{c_0r_\beta-1}{c_0r_\beta}}$ for $\beta$-mixing with $r_\beta\geq 1/c_0$. Then $\mathbb P(\mathcal B_{m,T})\geq 1-c_2\exp(-c_3m\gamma_m^2)-T^{-\left(\frac{1-c_0}{c_0}\right)}$ for some constants $c_2$ and $c_3$.
Moreover, on the event $\mathcal B_{m,T}$, for any $g\in B_{\mathcal H}(1)$ with $\|g\|_2\geq \gamma_m$, 
	\begin{equation}
	\frac{\|g\|_2}{2}\leq \|g\|_T\leq \frac{3}{2}\|g\|_2.
	\end{equation} 
\end{theorem}
The proofs for Theorems~\ref{theorem:Rademacher} and~\ref{thm:univariate} are provided in the appendix. Using Theorems \ref{theorem:Rademacher} and \ref{thm:univariate}, we are able to provide an upper bound on the noise term $\frac{1}{T}\sum_{t=1}^{T} \Delta_j(X_t)w_{t,j}$ in \eqref{eq:basicineq}. In particular, recalling that $\tilde{\gamma}_m=c_1\max\left\{\epsilon_m,\tilde{\epsilon}_m,\sqrt{\frac{\log(dT)}{m}}\right\}$, we have the following lemma.

\begin{lemma}\label{lemma:unibound}
Given $\tilde{\gamma}_m=\max(\gamma_m,\tilde{\epsilon}_m)$, on the event $\mathcal A_{m,T}\cap \mathcal B_{m,T}$, we have:
\begin{equation}
|\frac{1}{T}\sum_{t=1}^Tf_{j,k}(X_t)w_{t,j}|\leq 4\sqrt{2}\sqrt{\frac{m}{T}}(\tilde\gamma_m\| f_{j,k}\|_T +\tilde\gamma_m^2 \| f_{j,k}\|_{\mathcal H})
\end{equation} 
for any $f_{j,k}\in \mathcal H$, for all  $(j,k)\in\{1,2,...,d\}^2$.  
\end{lemma}

\subsection{Exploiting decomposability}
The reminder of our analysis involves conditioning on the event $\mathcal A_{m,T}\cap \mathcal B_{m,T}$.  Recalling the basic inequality \eqref{eq:basicineq} and using Lemma \ref{lemma:unibound}, on the event $\mathcal A_{m,T}\cap \mathcal B_{m,T}$ defined in Theorems~\ref{theorem:Rademacher} and ~\ref{thm:univariate}, we have:
\begin{equation*}
\frac{1}{2T}\sum_{t=1}^{T} B_Z(v_j+\hat{f}_j(X_t)||v_j+f^\ast_j(X_t))\leq 4\sqrt{2}\sqrt{\frac{m}{T}}\tilde\gamma_m\| \Delta_j\|_{T,1} +4\sqrt{2}\sqrt{\frac{m}{T}}\tilde\gamma_m^2 \| \Delta_j\|_{\mathcal H,1}+\lambda_T\| \Delta_j\|_{T,1} +\lambda_H \| \Delta_j\|_{\mathcal H,1}.
\end{equation*}
Recalling that $S_j$ denotes the true support of the unknown function $f^\ast_j$, we define $\Delta_{j,S_j}:=\sum_{k\in S_j}\Delta_{j,k}$, with a similar definition for $\Delta_{j,S_j^C}$.  We have that $\|\Delta_j\|_{T,1}=\|\Delta_{j,S_j}\|_{T,1}+\|\Delta_{j,S_j^C}\|_{T,1}$ with a similar decomposition for $\|\Delta_j\|_{\mathcal H,1}$. We are able to show that conditioned on event $\mathcal A_{m,T}\cap \mathcal B_{m,T}$, the quantities $\|\Delta_j\|_{\mathcal H,1}$ and $\|\Delta_j\|_{T,1}$ are not significantly larger than the corresponding norms as applied to the function $\Delta_{j,S_j}$. First, notice that we can obtain a sharper inequality in the process of getting our basic inequality \eqref{eq:basicineq}, that is,
\begin{equation*}
\frac{1}{2T}\sum_{t=1}^{T} B_Z(v_j+\hat{f}_j(X_t)||v_j+f^\ast_j(X_t))\leq \frac{1}{T}\sum_{t=1}^{T} \Delta_j(X_t)w_{t,j}+\lambda_T(\|f^\ast_j\|_{T,1}-\|f^\ast_j+\Delta_j\|_{T,1}) +\lambda_H (\|f^\ast_j\|_{\mathcal H,1}-\|f^\ast_j+\Delta_j\|_{\mathcal H,1}).
\end{equation*}
Using Lemma \ref{lemma:unibound} and the fact that Bregman divergence is non-negative, on event $\mathcal A_{m,T}\cap \mathcal B_{m,T}$ we obtain
\begin{equation*}
0\leq 4\sqrt{2}\sqrt{\frac{m}{T}}\tilde\gamma_m\| \Delta_j\|_{T,1} +4\sqrt{2}\sqrt{\frac{m}{T}}\tilde\gamma_m^2 \| \Delta_j\|_{\mathcal H,1}+\lambda_T(\|f^\ast_j\|_{T,1}-\|f^\ast_j+\Delta_j\|_{T,1}) +\lambda_H (\|f^\ast_j\|_{\mathcal H,1}-\|f^\ast_j+\Delta_j\|_{\mathcal H,1}).
\end{equation*}
Recall our choice $\lambda_T\geq 8\sqrt{2}\sqrt{\frac{m}{T}}\tilde\gamma_m, \lambda_H\geq 8\sqrt{2}\sqrt{\frac{m}{T}}\tilde\gamma_m^2$,
that yields
\begin{equation*}
0\leq \frac{\lambda_T}{2}\| \Delta_j\|_{T,1} +\frac{\lambda_H}{2}\| \Delta_j\|_{\mathcal H,1}+\lambda_T(\|f^\ast_j\|_{T,1}-\|f^\ast_j+\Delta_j\|_{T,1}) +\lambda_H (\|f^\ast_j\|_{\mathcal H,1}-\|f^\ast_j+\Delta_j\|_{\mathcal H,1}).
\end{equation*}
Now, for any $k\in S_j^C$, we have
\begin{equation*}
\|f^\ast_{j,k}\|_T-\|f^\ast_{j,k}+\Delta_{j,k}\|_T=-\|\Delta_{j,k}\|_T,\ and\ \|f^\ast_{j,k}\|_{\mathcal H}-\|f^\ast_{j,k}+\Delta_{j,k}\|_{\mathcal H}=-\|\Delta_{j,k}\|_{\mathcal H}.
\end{equation*}
On the other hand, for any $k\in S_j$, the triangle inequality yields
\begin{equation*}
\|f^\ast_{j,k}\|_T-\|f^\ast_{j,k}+\Delta_{j,k}\|_T\leq\|\Delta_{j,k}\|_T
\end{equation*}
with a similar inequality for the terms involving $\|\cdot\|_{\mathcal H}$. Given those bounds, we conclude that
\begin{equation}
0\leq \frac{\lambda_T}{2}\| \Delta_j\|_{T,1} +\frac{\lambda_H}{2}\| \Delta_j\|_{\mathcal H,1}+\lambda_T(\|\Delta_{j,S_j}\|_{T,1}-\|\Delta_{j,S_j^C}\|_{T,1}) +\lambda_H (\|\Delta_{j,S_j}\|_{\mathcal H,1}-\|\Delta_{j,S_j^C}\|_{\mathcal H,1}).
\end{equation}
Using the triangle inequality $\|\Delta_j\|\leq \|\Delta_{j,S_j}\|+\|\Delta_{j,S_j^C}\|$ for any norm and rearranging terms, we obtain
\begin{equation*}
\| \Delta_{j,S_j^C}\|_{T,1}+\| \Delta_{j,S_j^C}\|_{\mathcal H,1}\leq  3(\| \Delta_{j,S_j}\|_{T,1} +\| \Delta_{j,S_j}\|_{\mathcal H,1}),
\end{equation*}
which implies
\begin{equation}\label{eq:sparsitybound}
\| \Delta_j\|_{T,1}+\| \Delta_j\|_{\mathcal H,1}\leq  4(\| \Delta_{j,S_j}\|_{T,1} +\| \Delta_{j,S_j}\|_{\mathcal H,1}).
\end{equation}
This bound allows us to exploit the sparsity assumption, since in conjunction with Lemma \ref{lemma:unibound}, we have now bounded the right-hand side of the basic inequality \eqref{eq:basicineq} in terms involving only $\Delta_{j,S_j}$. In particular, still conditioning on event $\mathcal A_{m,T}\cap \mathcal B_{m,T}$ and applying \eqref{eq:sparsitybound}, we obtain
\begin{equation*}
\frac{1}{2T}\sum_{t=1}^{T} B_Z(v_j+\hat{f}_j(X_t)||v_j+f^\ast_j(X_t))\leq C\sqrt{\frac{m}{T}}\{\tilde{\gamma}_m\| \Delta_{j,S_j}\|_{T,1} +\tilde{\gamma}_m^2\| \Delta_{j,S_j}\|_{\mathcal H,1}\},
\end{equation*}
for some constant $C$, where we have recalled our choices $\lambda_T=O(\sqrt{\frac{m}{T}}\tilde{\gamma}_m)$ and $\lambda_H=O(\sqrt{\frac{m}{T}}\tilde{\gamma}_m^2)$. 
Finally, since both $\hat{f}_{j,k}$ and $f^\ast_{j,k}$ belong to $B_{\mathcal H}(1)$, we have
$$\|\Delta_{j,k}\|_{\mathcal H}\leq \|\hat{f}_{j,k}\|_{\mathcal H}+\|f^\ast_{j,k}\|_{\mathcal H}\leq 2,$$
which implies that $\|\Delta_{j,S_j}\|_{\mathcal H,1}\leq 2s_j$, and hence
\begin{equation*}
\frac{1}{2T}\sum_{t=1}^{T} B_Z(v_j+\hat{f}_j(X_t)||v_j+f^\ast_j(X_t))\leq C\sqrt{\frac{m}{T}}(\tilde{\gamma}_m\| \Delta_{j,S_j}\|_{T,1} +s_j\tilde{\gamma}_m^2).
\end{equation*}

\subsection{Exploiting strong convexity}

On the other hand, we are able to bound the Bregman divergence term on the left-hand side as well by noticing that \eqref{eq:sparsitybound} implies
\begin{equation}\label{eq:hatfbound}
\| \Delta_j\|_{\mathcal H,1}\leq  16s_j,
\end{equation}
since $\hat{f}_{j,k}$ and $f^\ast_{j,k}$ belong to $B_{\mathcal H}(1)$ with $\|\hat{f}_{j,k}\|_\infty\leq 1$ and $\|f^\ast_{j,k}\|_\infty\leq 1$.
Using bound \eqref{eq:hatfbound}, for any $t$, we conclude that
\begin{equation*}\begin{aligned}
|\hat{f}_j(X_t)|&=|\Delta_j(X_t)+f^\ast_j(X_t)|=|\sum_{k=1}^d \Delta_{j,k}(X_{t,k})+f^\ast_j(X_t)|\\
&\leq \sum_{k=1}^d \|\Delta_{j,k}\|_{\mathcal H}\max_k \sqrt{\mathcal K(X_{t,k},X_{t,k})}+|f^\ast_j(X_t)|\leq 16\sqrt{\sum_{i=1}^\infty\mu_i}s_j+|f^\ast_j(X_t)|\\
&\leq \left(16\sqrt{\sum_{i=1}^\infty\mu_i}+1\right)s_{\max}.
\end{aligned}\end{equation*}
Therefore, $v_j+\hat{f}_j(X_t),v_j+f^\ast_j(X_t)\in [v_{\min}-(16\sqrt{\sum_{i=1}^\infty\mu_i}+1)s_{\max},v_{\max}+(16\sqrt{\sum_{i=1}^\infty\mu_i}+1s_{\max}]$ where we have function $Z(\cdot)$ is $\vartheta$-strongly convex given Assumption~\ref{asp:strongconv}. Hence
\begin{equation}\label{eq:advancedbasic}
\frac{\vartheta}{2}\|\Delta_j\|_T\leq C\sqrt{\frac{m}{T}}\{\tilde{\gamma}_m\| \Delta_{j,S_j}\|_{T,1} +s_j\tilde{\gamma}_m^2\}.
\end{equation}

\subsection{Relating the $\mathcal L^2(\mathbb P_T)$ and $\mathcal L^2(\mathbb P)$ norms}
It remains to control the term $\|\Delta_{j,S_j}\|_{T,1}=\sum_{k\in S_j}\|\Delta_{j,k}\|_T$. Ideally we would like to upper bound it by $\sqrt{s_j}\|\Delta_{j,S_j}\|_T$. Such an upper bound would follow immediately if it were phrased in terms of the $\|\cdot\|_2$ rather than the $\|\cdot\|_T$ norm, but there are additional cross-terms with the empirical norm. Accordingly, we make use of two lemmas that relate the $\|\cdot\|_T$ norm and the population $\|\cdot\|_2$ norms for functions in $\mathcal F_j:=\cup_{S_j\subset \{1,2,...,d\}|S_j|=s_j}H_j(S_j)$.

In the statements of these results, we adopt the notation $g_j$ and $g_{j,k}$ (as opposed to $f_j$ and $f_{j,k}$) to be clear that our results apply to any $g_j\in \mathcal F_j$. We first provide an upper bound on the empirical norm $\|g_{j,k}\|_T$ in terms of the associated $\|g_{j,k}\|_2$ norm, one that holds uniformly over all components $k=1,2,...,d$.
\begin{lemma}\label{lemma:unidifference}
	On event $\mathcal B_{m,T}$,
	\begin{equation}
	\|g_{j,k}\|_T\leq 2\|g_{j,k}\|_2+\gamma_m \text{, for all } g_{j,k}\in B_{\mathcal H}(2)
	\end{equation}
	for any $(j,k)\in\{1,2,...,d\}^2$.
\end{lemma}
We now define the function class $2\mathcal F_j:=\{f+f'|f,f'\in \mathcal F_j\}$. Our second lemma guarantees that the empirical norm $\|\cdot\|_T$ of any function in $2\mathcal F_j$ is uniformly lower bounded by the norm $\|\cdot\|_2$.
\begin{lemma}\label{lemma:multibound}
	Given properties of $\gamma_m$ and $\delta^2_{m,j}=c_4\{\frac{s_j\log d}{m}+s_j\epsilon_m^2\}$, we define the event
		\begin{equation}
		\mathcal D_{m,T}=\{\forall j\in[1,2,...,d],\|g_j\|_T\geq \|g_j\|_2/2 \text{ for all } g_j\in 2\mathcal F_j\ with\ \|g_j\|_2\geq \delta_{m,j}\}
		\end{equation}
	where $m=T^{\frac{c_0r_\beta-1}{c_0r_\beta}}$ for $\beta$-mixing with $r_\beta\geq 1/c_0$. Then we have $\mathbb P(\mathcal D_{m,T})\geq1-c_2\exp(-c_3m(\min_j\delta_{m,j}^2))-T^{-\left(\frac{1-c_0}{c_0}\right)}$ where $c_2$, $c_3$ and $c_4$ are constants..
\end{lemma}
Note that while both results require bounds on the univariate function classes, they do not require global boundedness assumptions-that is, on quantities of the form $\|\sum_{k\in S_j}g_{j,k}\|_\infty$. Typically, we expect that the $\|\cdot\|_\infty$-norms of functions $g_j\in \mathcal F_j$ scale with $s_j$.

\subsection{Completing the proof}
Using Lemmas~\ref{lemma:unidifference} and ~\ref{lemma:multibound}, we complete the proof of the main theorem. For the reminder of the proof, let us condition on the events $\mathcal A_{m,T}\cap \mathcal B_{m,T}\cap \mathcal D_{m,T}$. Conditioning on the event $\mathcal B_{m,T}$, we have
\begin{equation}\label{eq:15}
\|\Delta_{j,S_j}\|_{T,1}=\sum_{k\in S_j}\|\Delta_{j,k}\|_T\leq 2\sum_{k\in S_j}\|\Delta_{j,k}\|_2+s_j\gamma_m\leq 2\sqrt{s_j}\|\Delta_{j,S_j}\|_2+s_j\gamma_m.
\end{equation}
Our next step is to bound $\|\Delta_{j,S_j}\|_2$ in terms of $\|\Delta_{j,S_j}\|_T$ and $s_j\gamma_m$. We split our analysis into two cases.\\
{\bf Case 1:} If $\|\Delta_{j,S_j}\|_2<\delta_{m,j}=\Theta(\sqrt{s_j}\gamma_m)$, then we conclude that $\|\Delta_{j,S_j}\|_{1,T}\leq Cs_j\gamma_m$.\\
{\bf Case 2:} Otherwise, we have $\|\Delta_{j,S_j}\|_2\geq \delta_{m,j}$. Note that the function $\Delta_{j,S_j}=\sum_{k\in S_j}\Delta_{j,k}$ belongs to the class $2\mathcal F_j$
so that it is covered by the event $\mathcal D_{m,T}$. In particular, conditioned on the event $\mathcal D_{m,T}$, we have $\|\Delta_{j,S_j}\|_2\leq 2\|\Delta_{j,S_j}\|_T$. Combined with the previous bound~\eqref{eq:15}, we conclude that 
$$\|\Delta_{j,S_j}\|_{T,1}\leq C\{\sqrt{s_j}\|\Delta_{j,S_j}\|_{T,2}+s_j\gamma_m\}.$$
Therefore in either case, a bound of the form $\|\Delta_{j,S_j}\|_{T,1}\leq C\{\sqrt{s_j}\|\Delta_{j,S_j}\|_{T,2}+s_j\gamma_m\}$ holds. Substituting the inequality in the bound \eqref{eq:advancedbasic} yields
\begin{equation*}
\frac{\vartheta}{2}\| \Delta_j\|^2_T\leq C_1\sqrt{\frac{m}{T}}(\sqrt{s_j}\tilde\gamma_m\lVert \Delta_{j,S_j}\rVert_T +s_j\tilde\gamma_m^2).
\end{equation*}
The term $\|\Delta_{j,S_j}\|_T$ on the right side of the inequality is bounded by $\|\Delta_j\|_T$ and the inequality still holds after replacing $\|\Delta_{j,S_j}\|_T$ by $\|\Delta_j\|_T$. Through rearranging terms in that inequality, we get,
\begin{equation}
\|\Delta_j\|_T^2 \leq 2C_1\frac{1}{\vartheta}\sqrt{\frac{m}{T}}\left(\sqrt{s_j}\tilde\gamma_m\| \Delta_j\|_T+s_j\tilde\gamma_m^2\right).
\end{equation}
Because $\frac{m}{T}\leq 1$ and $\frac{1}{\vartheta}\geq 1$, we can relax the inequality to
\begin{equation}
\|\Delta_j\|_T^2 \leq 2C_1\left(\frac{1}{\vartheta}\left(\frac{m}{T}\right)^{1/4}\sqrt{s_j}\tilde\gamma_m\| \Delta_j\|_T+\frac{1}{\vartheta^2}\left(\frac{m}{T}\right)^{1/2}s_j\tilde\gamma_m^2\right).
\end{equation}
We can derive a bound on $\|\Delta_j\|_T$ from that inequality, which is
\begin{equation}\begin{aligned}
\|\Delta_j\|_T^2&\leq C_2\frac{s_j}{\vartheta^2}\sqrt{\frac{m}{T}}\tilde{\gamma}_m^2=C_2\frac{s_j}{\vartheta^2}\sqrt{\frac{m}{T}}\left(\frac{\log (dT)}{m}+\max(\epsilon_m,\tilde{\epsilon}_m)^2\right)\\
&=C_2\frac{s_j}{\vartheta^2}\left(\frac{\log (dT)}{\sqrt{mT}}+\sqrt{\frac{m}{T}}\max(\epsilon_m,\tilde{\epsilon}_m)^2\right)\\
&=C_2\frac{s_j}{\vartheta^2}\left(\frac{\log (dT)}{\sqrt{mT}}+\sqrt{\frac{m}{T}}\tilde{\epsilon}_m^2\right),
\end{aligned}\end{equation}
where $C_2$ only depends on $C_1$. That completes the proof.
\section{Numerical experiments}\label{SecSimualtions}
Our experiments are two-fold. First we perform simulations that validate the theoretical results in Section~\ref{SecMain}. We then apply the SpAM framework on a Chicago crime dataset and show its improvement in prediction error and ability to discover additional interesting patterns beyond the parametric model. Instead of using the sparse and smooth objective in this paper, we implement a computationally faster approach through the R CRAN package ``SAM'', which includes the first penalty term $\|f_j\|_{1,T}$ but not the second term $\|f_j\|_{1,\mathcal H}$~(\cite{zhao12}). We also implemented our original optimization problem in `cvx' however due to computational challenges this does not scale. Hence we use ``SAM''. 

\subsection{Simulations}
We validate our theoretical results with experimental results performed on synthetic data. We generate many trials with known underlying parameters and then compare the estimated function values with the true values. For all trials the constant offset vector $v$ is set identically at $0$. Given an initial vector $X_0$, samples are generated consecutively using the equation $X_{t+1,j} = f^\ast_j(X_t)+w_{t+1,j}$, where $w_{t+1,j}$ is the noise chosen from a uniform distribution on the interval $[-0.4,0.4]$ and $f^\ast_j$ is the signal function, which means that the log-partition function $Z(.)$ is the standard quadratic $Z(x) = \frac{1}{2}x^2$ and the sufficient statistic $\varphi(x) = x$. The signal function $f^\ast_j$ is assigned in two steps to ensure that the Markov chain mixes and we incorporate sparsity. In the first step, we define sparsity parameters $\{s_j\}_{j=1}^d$ all to be $3$ (for convenience) and set up a $d$ by $d$ sparse matrix $A^\ast$, which has $3$ non-zero off-diagonal values on each row drawn from a uniform distribution on the interval $[-\frac{1}{2s},\frac{1}{2s}]$ and all $1$ on diagonals. In the second step, given a polynomial order parameter $r$, we map each value $X_{t,k}$ in vector $X_t$ to $(\Phi_1(X_{t,k}),\Phi_2(X_{t,k}),...,\Phi_r(X_{t,k}))$ in $\mathbb R^r$ space, where $\Phi_i(x)=\frac{x^i}{i!}$ for any $i$ in $\{1,2,..,r\}$. We then randomly generate standardized vectors $(b^1_{j,k},b^2_{j,k},b^3_{j,k})$ for every $(j,k)$ in $\{1,2,...,d\}^2$ and define $f^\ast_j$ as $f^\ast_j(X_t) = \sum_{k=1}^d A^*_{j,k}(\sum_{i=1}^r b^i_{j,k}\Phi_i(X_{t,k}))$. The tuning parameter $\lambda_T$ is chosen to be $3\sqrt{\log(dr)/T}$ following the theory. We focus on polynomial kernels for which we have theoretical guarantees in Lemma~\ref{example:1} and Corollary~\ref{Cor:1} since the ``SAM'' package is suited to polynomial basis functions.

The simulation is repeated $100$ times with $5$ different values of $d$ ($d=8,16,32,64,128$), $5$ different numbers of time points ($T=80,120,160,200,240$), and $3$ different polynomial order parameters ($r=1,2,3$) for each repetition. These design choices are made to ensure the sequence $(X_t)_{t=0}^T$ is stable and mixes. Other experimental settings were also run with similar results.

\begin{figure}[t]
    \begin{subfigure}[t]{0.5\textwidth}
        \centering
	    \includegraphics[width=0.8\linewidth]{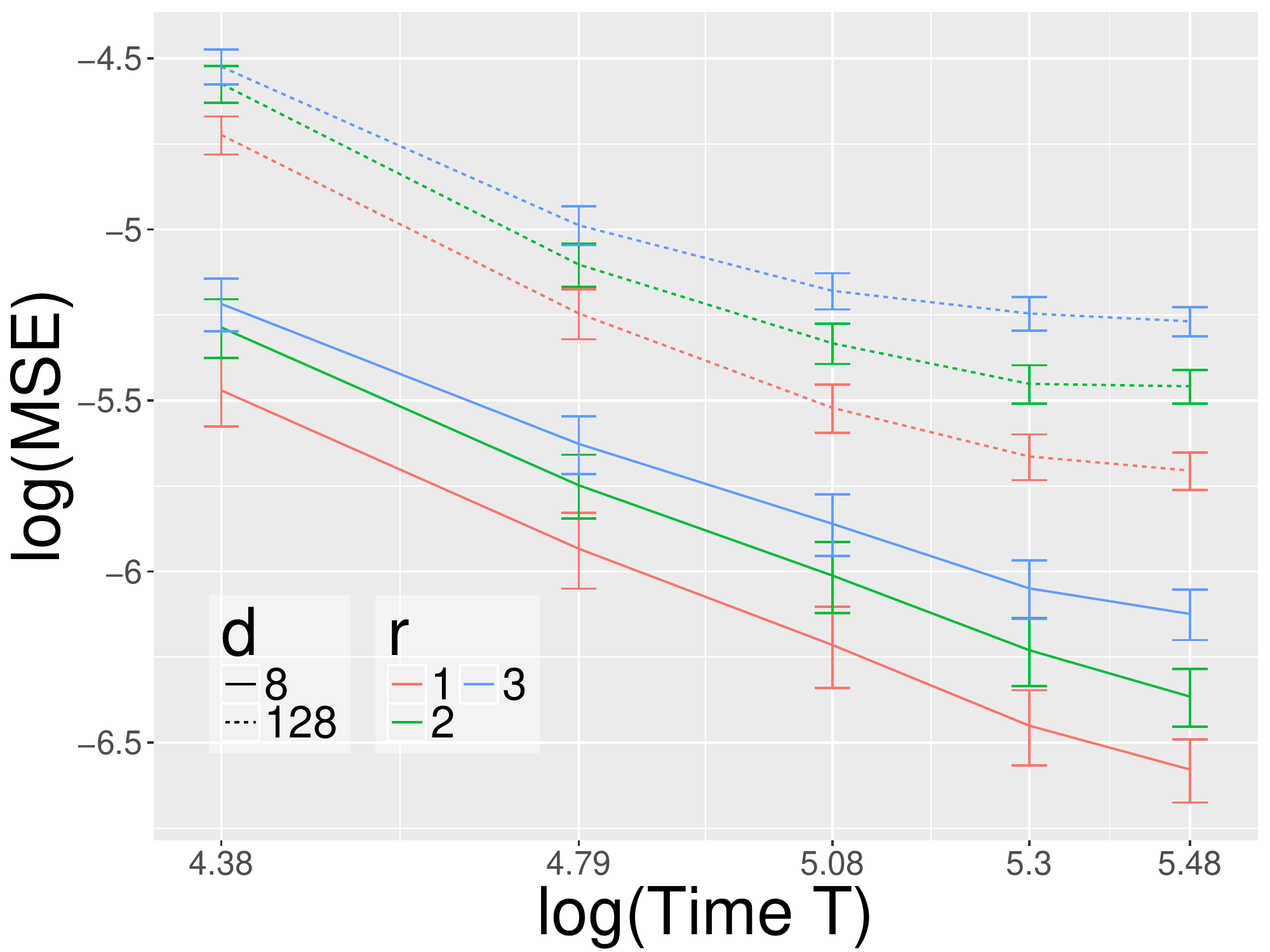}
	    \caption{log(MSE) vs. log(Time T)}
	\end{subfigure}
	\begin{subfigure}[t]{0.5\textwidth}
	    \centering
	    \includegraphics[width=0.8\linewidth]{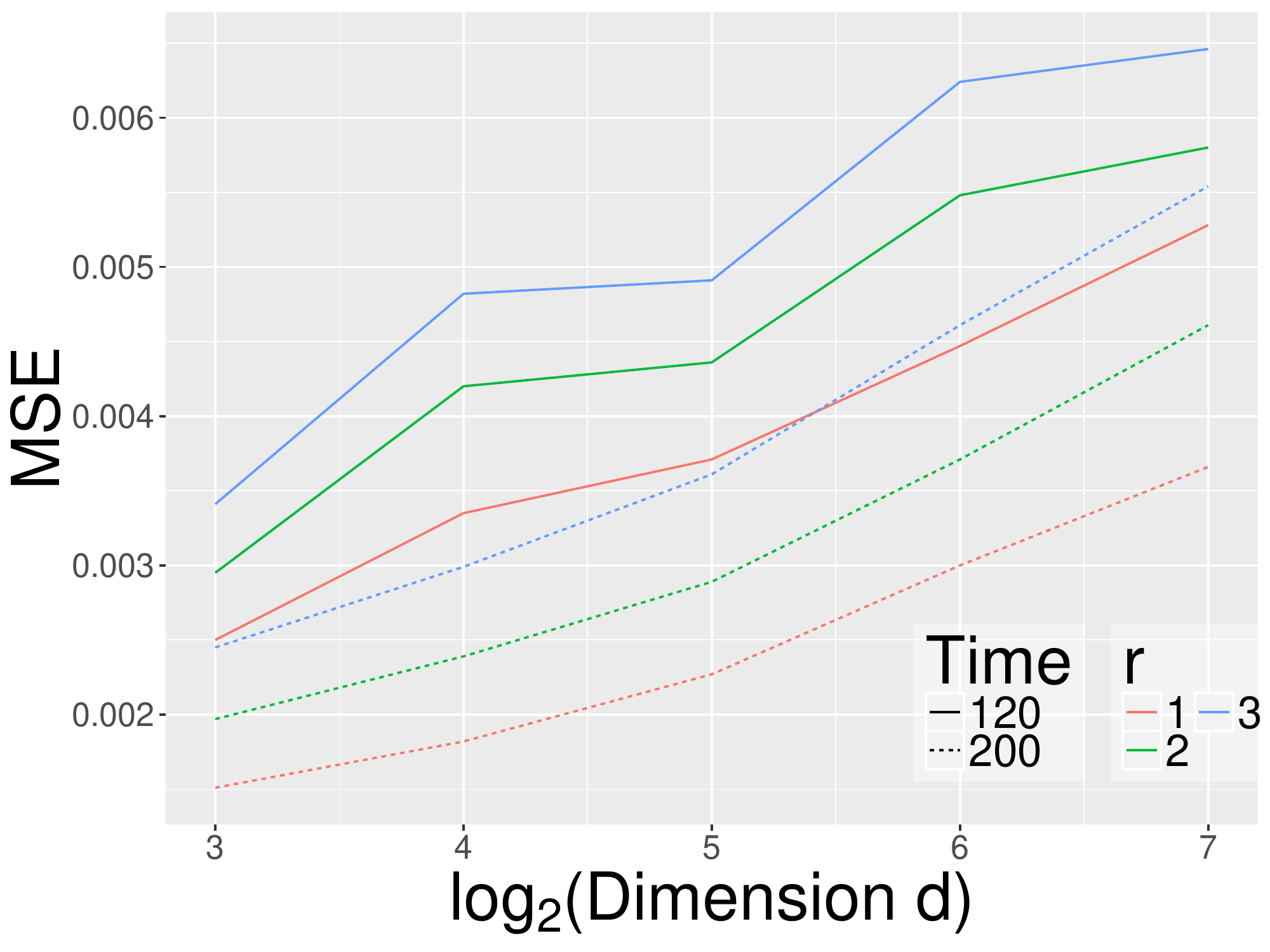}
	    \caption{MSE vs. log(Dimension d)}
	\end{subfigure}
	\caption{(a) shows the logarithm of MSE over a range of $\log T$ values, from 80 to 240 under the regression setting. (b) shows the MSE over a range of $\log d$ values, from 8 to 128 under the regression setting. In all plots the mean value of 100 trials is shown, with error bars denoting the $90\%$ confidence interval for plot (a). For plot (b), we also have error bars results but we do not show them for the cleanness of the plot.}
    \label{fig:sim}
\end{figure}

We present the mean squared error (MSE) of our estimates in Fig. \ref{fig:sim}. Since we select $r$ values from the same vector $(b^1_{j,k},b^2_{j,k},b^3_{j,k})$ for all polynomial order parameters, the MSE for different $r$ is comparable and will be higher for larger $r$ because of stronger absolute signal value. 
In Fig. \ref{fig:sim}(a), we see that MSE decreases in the rate between $T^{-1}$ and $T^{-0.5}$ for all combinations of $r$ and $d$. For larger $d$, MSE is larger and the rate becomes slower. In Fig. \ref{fig:sim}(b), we see that MSE increases slightly faster than the $\log d$ rate for all combinations of $r$ and $T$ which is consistent with Theorem~\ref{thm:key} and Corollary 1.

\begin{figure}[t]
    \begin{subfigure}[t]{0.5\textwidth}
        \centering
	    \includegraphics[width=0.8\linewidth]{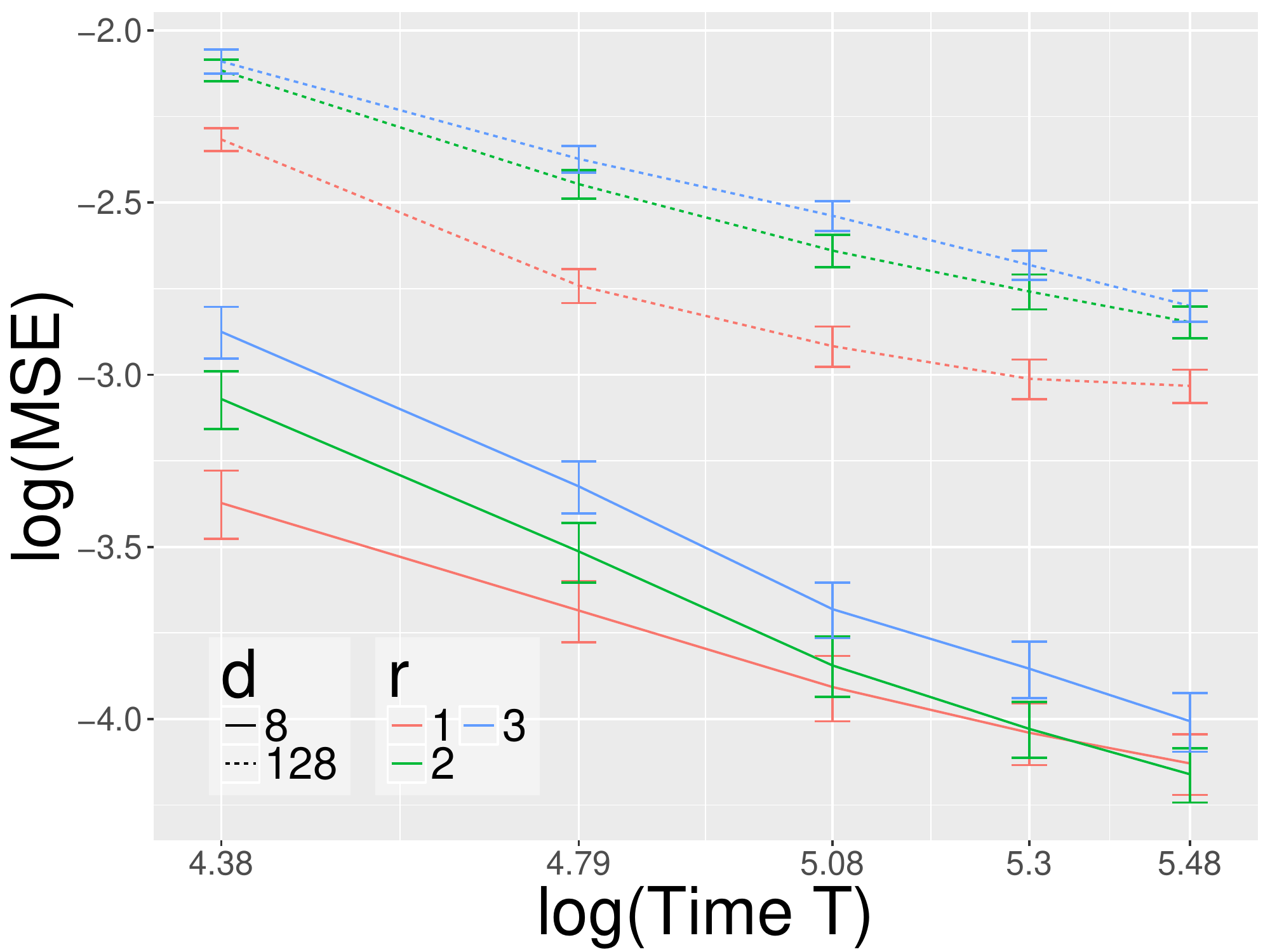}
	    \caption{log(MSE) vs. log(Time T)}
	\end{subfigure}
	\begin{subfigure}[t]{0.5\textwidth}
	    \centering
	    \includegraphics[width=0.8\linewidth]{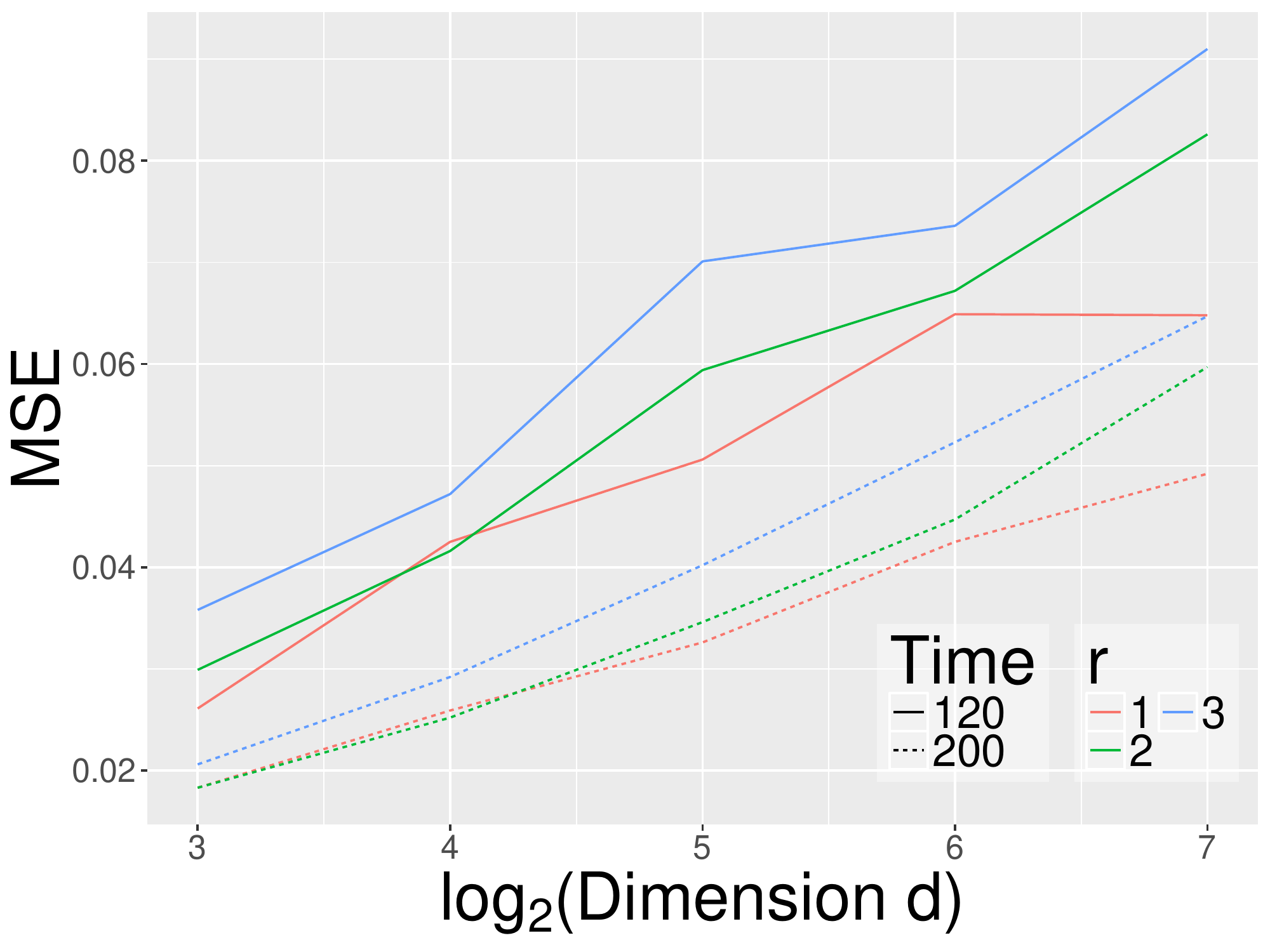}
	    \caption{MSE vs. log(Dimension d)}
	\end{subfigure}
	\caption{(a) shows the logarithm of MSE behavior over a range of $\log T$ values, from 80 to 240 for the Poisson process. (b) shows the MSE behavior over a range of $\log d$ values, from 8 to 128 for the Poisson process. In all plots the mean value of 100 trials is shown, with error bars denoting the $90\%$ confidence interval for plot (a). For plot (b), we also have error bars results but we do not show them for the cleanness of the plot.}
    \label{fig:simpoi}
\end{figure}

Similarly we consider the Poisson link function and Poisson process for modeling count data. Given an initial vector $X_0$, samples are generated consecutively using the equation $X_{t+1,j} \sim \mbox{Poisson}(\exp(f^\ast_j(X_t)))$, where $f^\ast_j$ is the signal function. The signal function $f^\ast_j$ is again assigned in two steps to ensure the Poisson Markov process mixes. In the first step, we define sparsity parameters $\{s_j\}_{j=1}^d$ all to be $3$ and set up a $d$ by $d$ sparse matrix $A^\ast$, which has $3$ non-zero values on each row set to be $-2$ (this choice ensures the process mixes). In the second step given a polynomial order parameter $r$, we map each value $X_{t,k}$ in vector $X_t$ to $(\Phi_1(X_{t,k}),\Phi_2(X_{t,k}),...,\Phi_r(X_{t,k}))$ in $\mathbb R^r$, where $\Phi_i(x)=\frac{x^i}{i!}$ for any $i$ in $\{1,2,..,r\}$. We then randomly generate standardized vectors $(b^1_{j,k},b^2_{j,k},b^3_{j,k})$ for every $(j,k)$ in $\{1,2,...,d\}^2$ and define $f^\ast_j$ as $f^\ast_j(X_t) = \sum_{k=1}^d A^*_{j,k}(\sum_{i=1}^r b^i_{j,k}\Phi_i(X_{t,k}))$. The tuning parameter $\lambda_T$ is chosen to be $1.3(\log d\log T)(\sqrt{r}/\sqrt{T})$. The simulation is repeated $100$ times with $5$ different numbers of time series ($d=8,16,32,64,128$), $5$ different numbers of time points ($T=80,120,160,200,240$) and $3$ different polynomial order parameters ($r=1,2,3$) for each repetition. These design choices are made to ensure the sequence $(X_t)_{t=0}^T$ mixes. Other experimental settings were also considered with similar results, but are not included due to space constraints.

We present the mean squared error (MSE) of our estimations in Fig. \ref{fig:simpoi}. Since we select $r$ values from the same vector $(b^1_{j,k},b^2_{j,k},b^3_{j,k})$ for all polynomial order parameters, the MSE tends to be higher for larger $r$ because the process has larger variance. 
In Fig. \ref{fig:simpoi} (a), we see that MSE decreases in the rate between $T^{-1}$ and $T^{-0.5}$ for all combinations of $r$ and $d$. For larger $d$, MSE is larger and the rate becomes slower. In Fig. \ref{fig:simpoi} (b), we see that MSE increases slightly faster than the $log(d)$ rate for all combinations of $r$ and $T$ which is consistent with our theory.

\subsection{Chicago crime data}

We now evaluate the performance of the SpAM framework on a Chicago crime dataset to model incidents of severe crime in different community areas of Chicago.
\footnote{This dataset reflects reported incidents of crime that occurred in the City of Chicago from 2001 to present. Data is extracted from the Chicago Police Department's CLEAR (Citizen Law Enforcement Analysis and Reporting) system \url{https://data.cityofchicago.org}}.
We are interested in predicting the number of homicide and battery (severe crime) events every two days for 76 community areas over a two month period. The recorded time period is April 15, 2012 to April 14, 2014 as our training set and we choose the data from April 15, 2014 to June 14, 2014 to be our test data. In other words, we consider dimension $d = 76$ and time range $T = 365$ for training set and $T = 30$ for the test set. Though the dataset has records from 2001, we do not use all previous data to be our training set since we do not have stationarity over a longer period. We choose a 2 month test set for the same reason. We choose time horizon to be two days so that number of crimes is counted over each two days. Since we are modeling counts, we use the Poisson GLM and the exponential link $Z(x) = e^x$.

\begin{figure}[ht]
    \centering
    \includegraphics[width=0.8\linewidth]{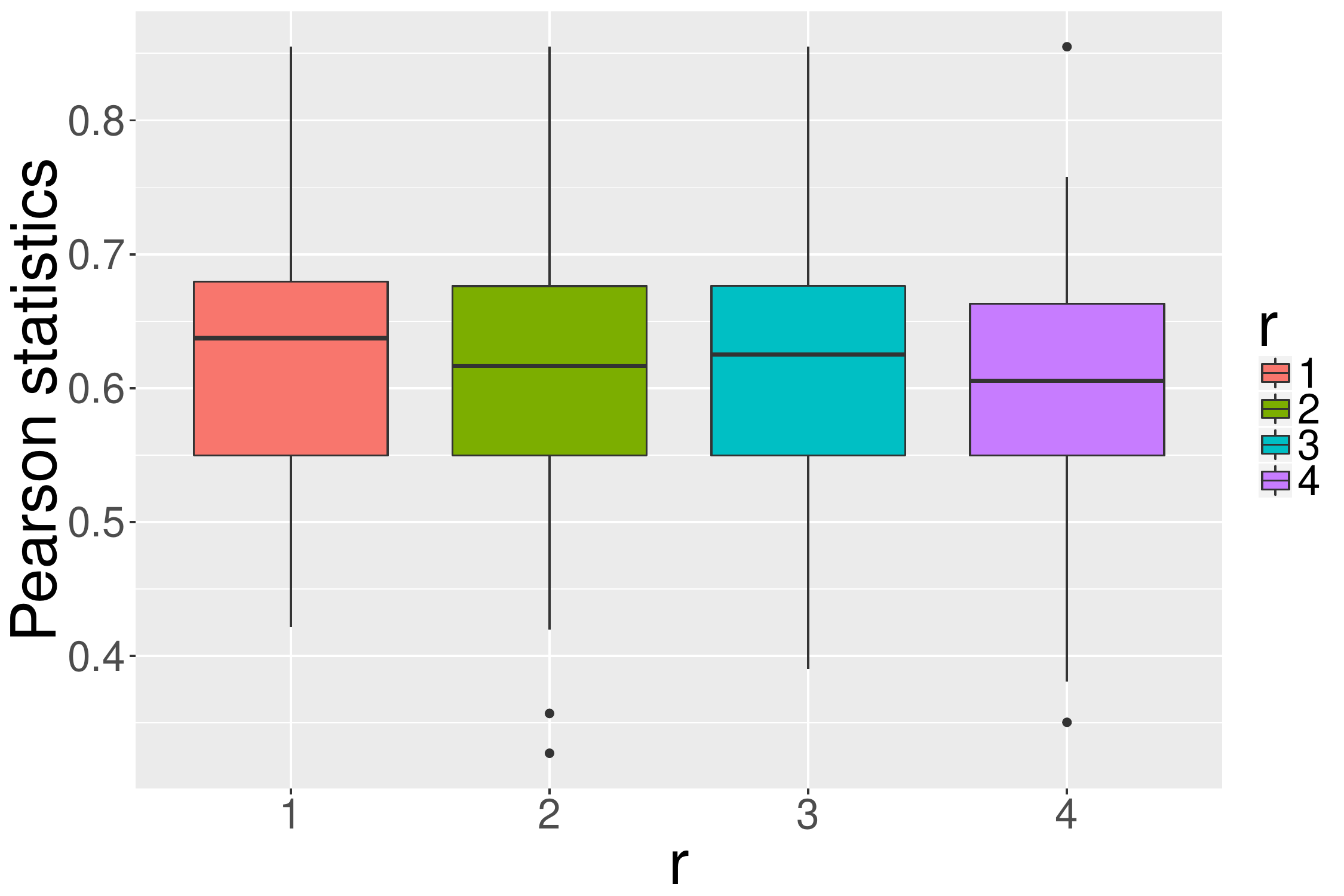}
    \caption{The boxplot shows the performance of SpAM on crime data measured by Pearson statistic for $r=1,2,3,4$-degrees of freedom in B-spline basis.}
    \label{fig:cvloss}
\end{figure}

\begin{figure}[!ht]
    \begin{subfigure}[t]{0.49\linewidth}
        \centering
	    \includegraphics[width=0.98\linewidth]{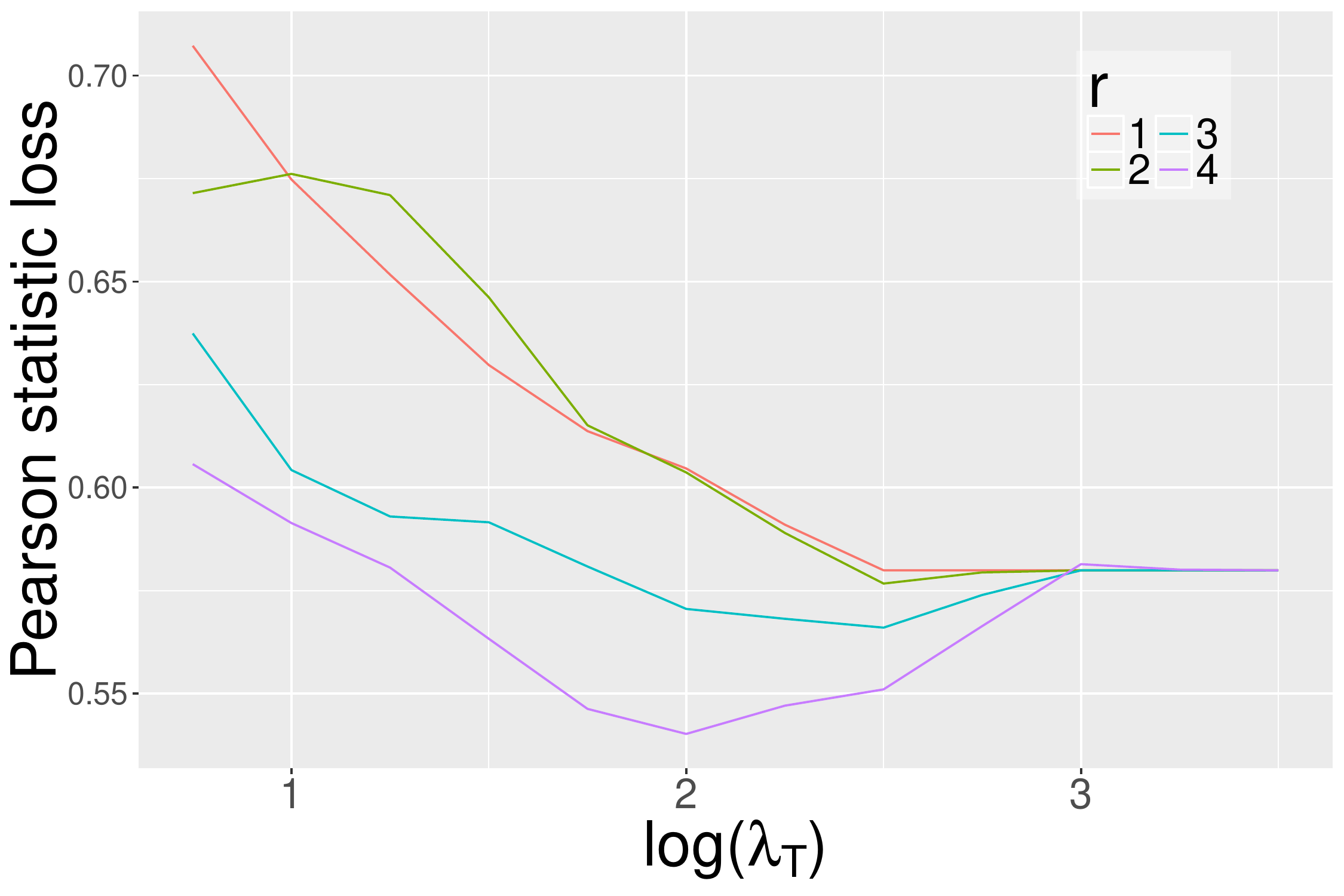}
	    \caption{Performance on community area 34}
	\end{subfigure}
	\begin{subfigure}[t]{0.49\linewidth}
	    \centering
	    \includegraphics[width=0.98\linewidth]{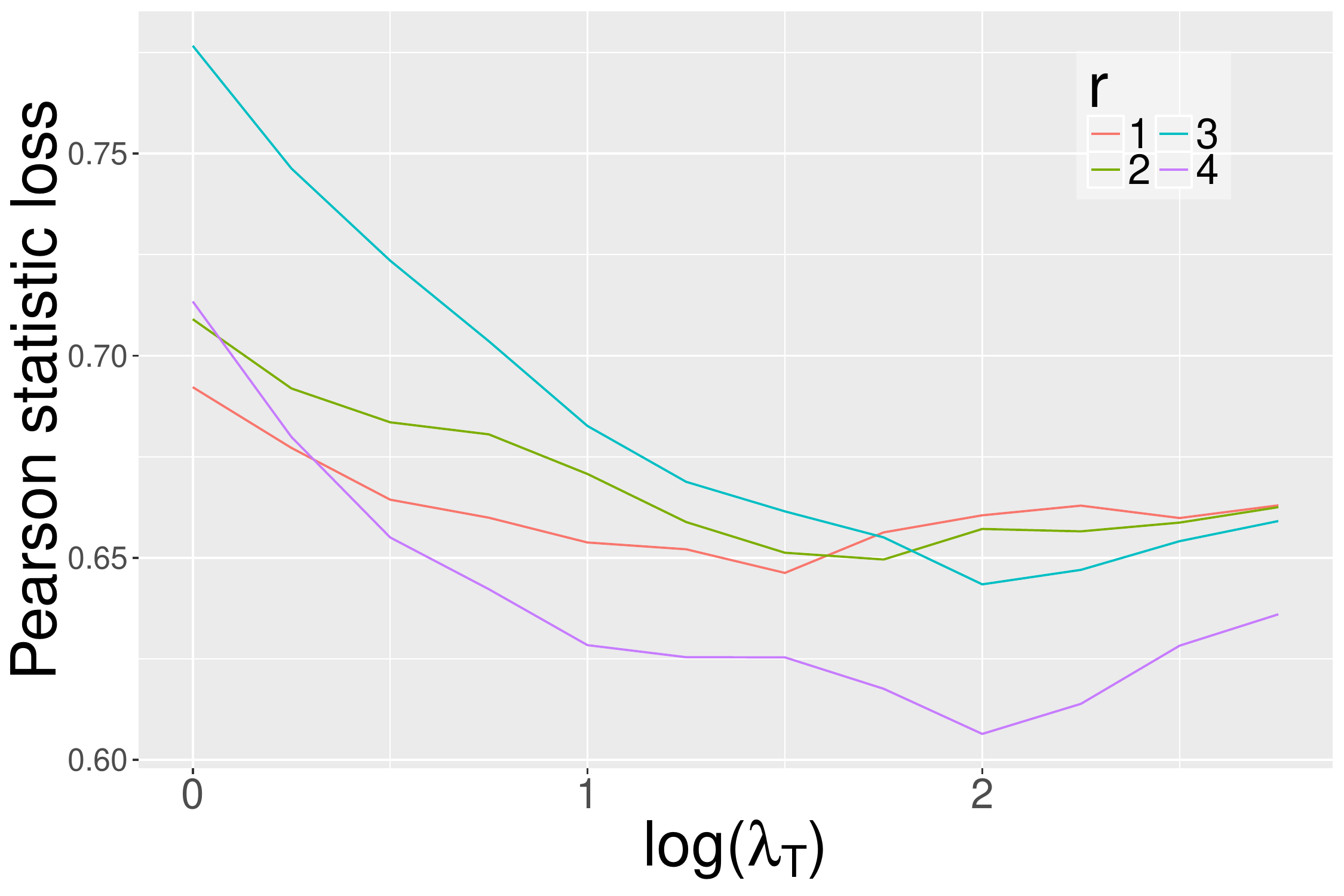}
	    \caption{Performance on community area 56}
	\end{subfigure}
	\caption{ (a) shows the Pearson statistic loss on the number of crimes in community area 34. (b) shows the Pearson statistic loss on the number of crimes in community area 56.
    }
    \label{fig:path}
\end{figure}

We apply the ``SAM'' package for this task using B-spline as our basis. The degrees of freedom $r$ are set to $1,2,3$ or $4$, where $1$ means that we only use linear basis. In the first part of the experiment, we choose the tuning parameter $\lambda_T$ using 3-cross validation; the validation pairs are chosen as $60$ days back (i.e., February 15, 2012 to February 14, 2014 as the training set and February 15, 2014 to April 14, 2014 as the testing set), $120$ days back and $180$ days back from April 15, 2012 and April 15, 2014 but with the same time range as the training set and test set. Then we test SpAM with this choice of $\lambda_T$. The performance of the model is measured by Pearson chi-square statistic, which is defined as 
\begin{equation*}
    \frac{1}{30}\sum_{t=0}^{29}\frac{(X_{t+1,j}-\hat{f}_j(X_t))^2}{\hat{f}_j(X_t)}
\end{equation*}
on the $30$ test points for the $j^{th}$ community area. The Pearson chi-square statistic is commonly used as the goodness-of-fit measure for discrete observations~\cite{hosmer97}. In Fig. \ref{fig:cvloss}, we show a box plot for the test loss on 17 non-trivial community areas, where ``trivial'' means that the number of crimes in the area follows a Poisson distribution with constant rate, which tells us that there is no relation between that area and other areas and no relation between different time. From Fig. \ref{fig:cvloss}, we can see that as basis become more complex from linear to B-spline with 4 degrees of freedom, the performance of fitting is gradually (although not majorly) improved. The main benefit of using higher-order (non-parametric) basis is revealed in Fig.~\ref{fig:path} where we pick two community areas and plot the $\lambda_T$ path performance for every $r$ in Fig.~\ref{fig:path}. 
In the examples of two community areas shown in Fig. \ref{fig:path}, we can see that the non-parametric SpAM has a lower test loss than linear model ($r=1$). For community area 34, when $r$ is set to be 3 and 4, the SpAM model discovers meaningful influences of other community areas on that area while the model with $r$ equal to 1 or 2 choose a constant Poisson process as the best fitting. A similar conclusion holds for community area 56. Here $r=1$ corresponds to the parametric model in~\cite{hall16}.

\begin{figure}[ht]
    \begin{subfigure}[t]{0.49\linewidth}
        \centering
	    \includegraphics[width=0.98\linewidth]{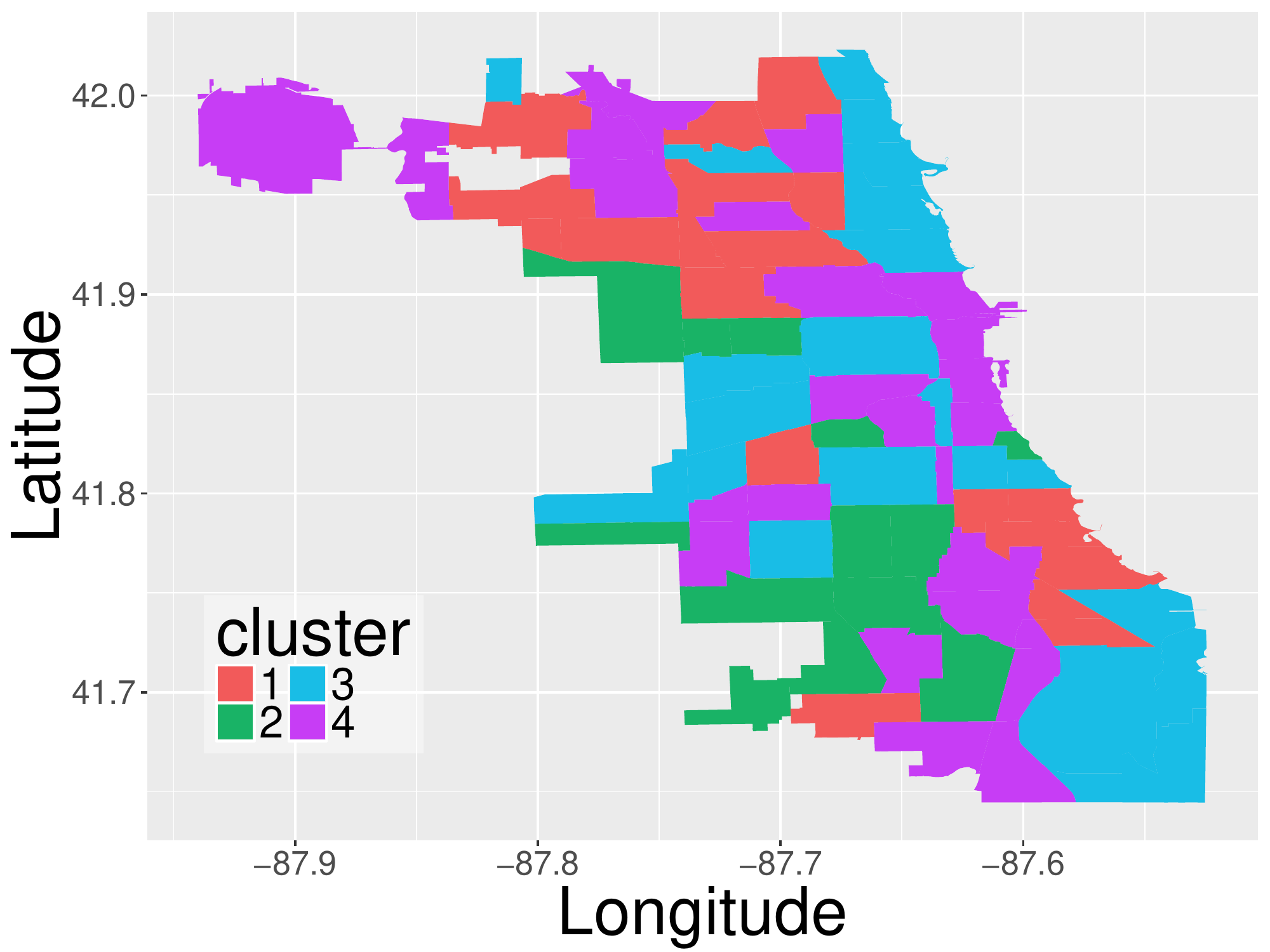}
	    \caption{$r = 3$}
	\end{subfigure}
	\begin{subfigure}[t]{0.49\linewidth}
	    \centering
	    \includegraphics[width=0.98\linewidth]{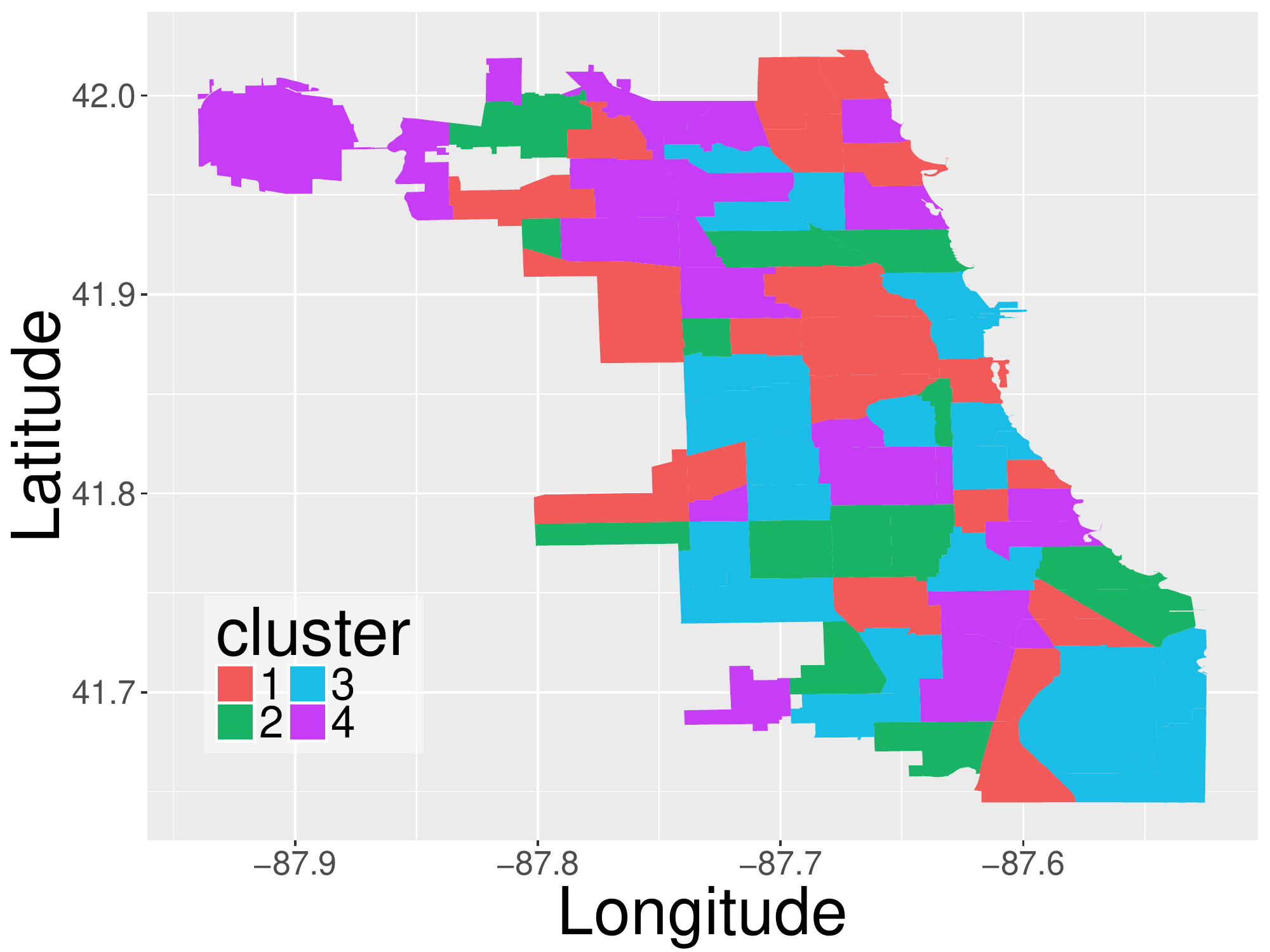}
	    \caption{$r = 1$}
	\end{subfigure}
    \begin{subfigure}[t]{0.49\linewidth}
        \centering
	    \includegraphics[width=0.98\linewidth]{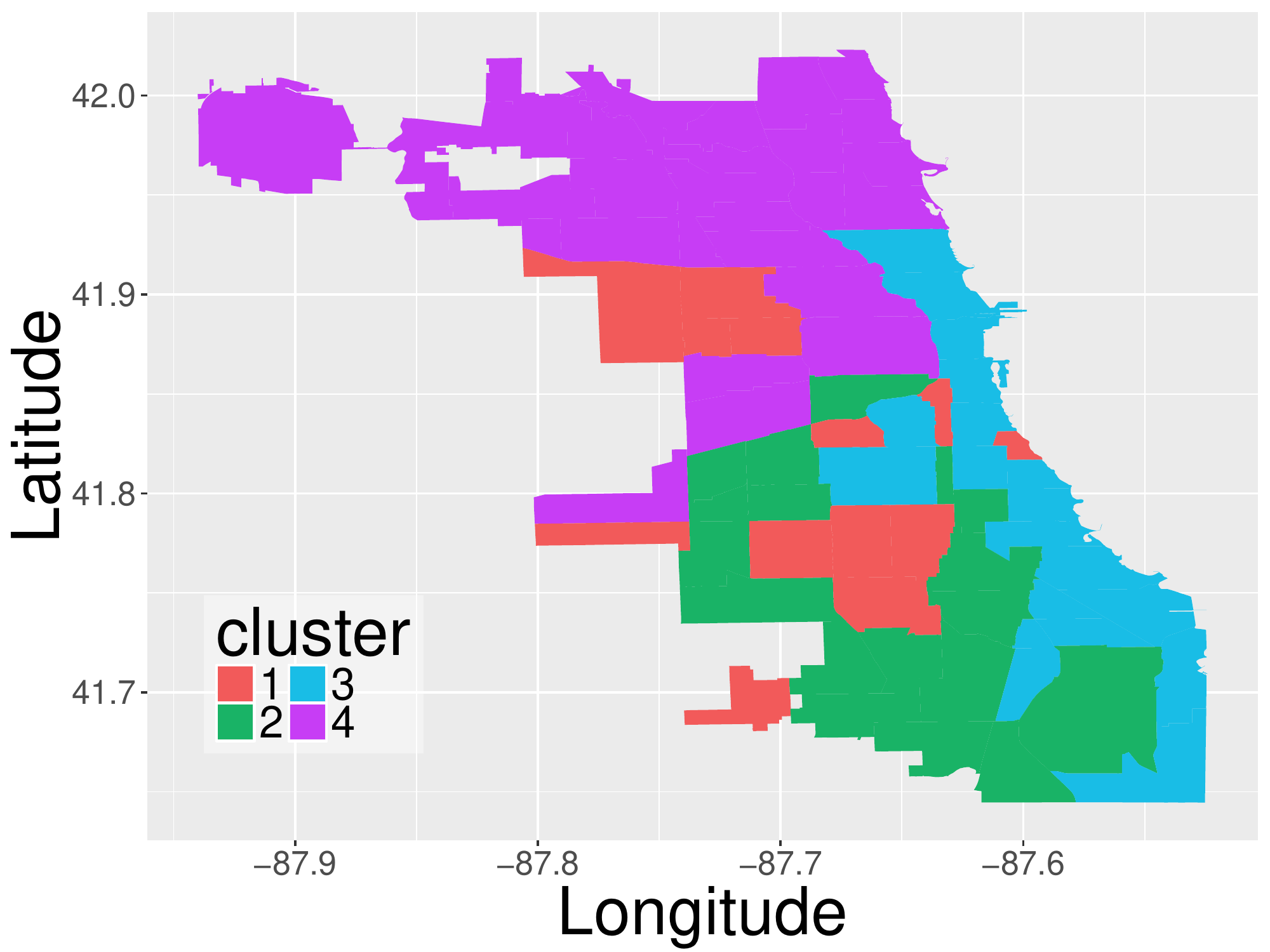}
	    \caption{$r = 3$ using location information}
	\end{subfigure}
	\begin{subfigure}[t]{0.49\linewidth}
	    \centering
	    \includegraphics[width=0.98\linewidth]{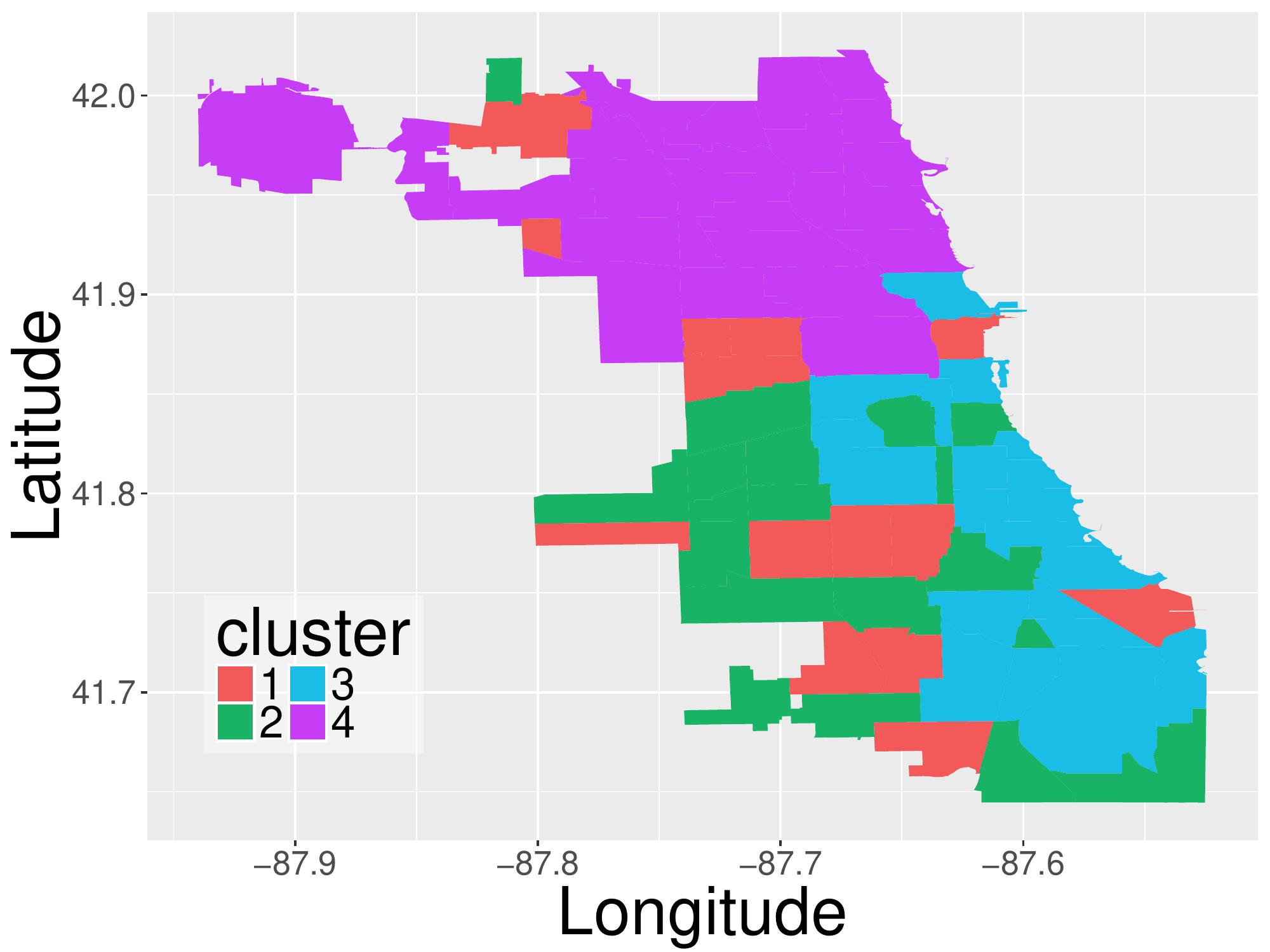}
	    \caption{$r = 1$ using location information}
	\end{subfigure}
	\caption{(a) shows the clusters given by spectral clustering using the adjacency matrix from SpAM with polynomial order $r = 3$, (b) shows the clusters when polynomial order $r = 1$. To derive clusters in (c), compared to (a), we add location information to the adjacency matrix for $r = 3$. Similarly, we get clusters in (d) using location information compared to (b) for $r = 1$.
    }
    \label{fig:cluster}
\end{figure}

Finally, we present a visualization of the estimated network for the Chicago crime data. Since the estimated model is a network, the sparse patterns can be represented as an adjacency matrix where $1$ in the $i^{th}$ row and $j^{th}$ column means that the $i^{th}$ community area has influence on the $j^{th}$ community area and $0$ means no effect. Given the adjacency matrix, we can use spectral clustering to generate clusters for different polynomial order $r$'s used in SpAM model, which are shown in Figs. \ref{fig:cluster} (a) and (b). For each case, even the location information is not used in learning at all, we find that the close community areas are clustered together. We see that the patterns from the non-parametric model ($r=3$) is different from the parametric generalized linear model ($r=1$) and they seem more smooth. It tells us that the non-parametric model proposed in this work can help us to discover additional patterns beyond the linear model. Even in other tasks, the clusters cannot represent the location information very well. In~\cite{binkiewicz17, zhang17}, the authors proposed a covariate-assisted method to deal with this problem, which applies spectral clustering on $L + \lambda X^TX$, where $L$ is the adjacency matrix, $X$ are the covariates (latitude and longitude in our case), and $\lambda$ is a tuning parameter. By using location information as the assisted covariate in spectral clustering, we obtain results in Fig. \ref{fig:cluster} (c)(d). Since the location information is used, we see in both cases that community areas are almost clustered in four groups based on location information. Again, we find that the patterns from non-parametric model is different from the linear model and the separation between clusters is slightly clearer.

\section{Acknowledgement}
This work is partially supported by NSF-DMS 1407028, ARO W911NF-17-1-0357, NGA HM0476-17-1-2003, NSF-DMS 1308877.
\clearpage

\bibliographystyle{plain}
\bibliography{PaperSPAMTimeSeries}

\clearpage

\section{Appendix}
In this Appendix, we give the proofs for Theorem~\ref{thm:follow}, the two examples in Subsection~\ref{section:twoexamples}, Theorems~\ref{theorem:Rademacher} and Theorem~\ref{thm:univariate} (which are the key results used in the proof of Theorem~\ref{thm:key}). Then proofs for other Lemmas are presented in Subsection~\ref{otherproofs}. 

\subsection{Proof of Theorem~\ref{thm:follow}}
The outline of this proof is the same as the outline of the proof for Theorem~\ref{thm:key}. The key difference here is that, given a $\phi$-mixing process with $0.781\leq r_\phi\leq 2$, we are able to derive sharper rates for Theorem~\ref{thm:univariate} and Lemma~\ref{lemma:multibound}, which result in $m=T^{\frac{r_\phi}{r_\phi+2}}$. For $r_\phi\leq 2$ this rate is sharper since $T^{\frac{r_\phi}{r_\phi+2}}\geq T^{\frac{r_\phi-1}{r_\phi}}$. Specifically, using the concentration inequality from~\cite{kontorovich08}, we show two Lemmas which give us a larger $m$ than Theorem~\ref{thm:univariate} and Lemma~\ref{lemma:multibound}.
\begin{lemma}\label{lemma:newunivariate}
	Define the event
	\begin{equation*}
		\mathcal B_{m,T}=\left\{\sup_{j,k}\sup_{f_{j,k}\in B_H(1), \|f_{j,k}\|_2\leq \gamma_m} |\|f_{j,k}\|_T-\|f_{j,k}\|_2|\leq \frac{\gamma_m}{2}\right\}.
	\end{equation*}

For a stationary $\phi$-mixing process $(X_t)_{t=0}^T$ with $0.781 \leq r_\phi\leq 2$ and $m=T^{\frac{r_\phi}{r_\phi+2}}$, we have $\mathbb P(\mathcal B_{m,T})\geq1-c_2\exp(-c_3(m\gamma_m^2)^2)$ where $c_2$ and $c_3$ are constants.

Moreover, on the event $\mathcal B_{m,T}$, for any $g\in B_{\mathcal H}(1)$ with $\|g\|_2\geq \gamma_m$, 
	\begin{equation}
	\frac{\|g\|_2}{2}\leq \|g\|_T\leq \frac{3}{2}\|g\|_2.
	\end{equation} 
\end{lemma}

\begin{lemma}\label{lemma:newmultibound}
	Given properties of $\gamma_m$ and $\delta^2_{m,j}=c_4\{\frac{s_j\log(d)}{m}+s_j\epsilon_m^2\}$, we define the event
		\begin{equation*}
		\mathcal D_{m,T}=\{\forall j\in[1,2,...,d],\|g_j\|_T\geq \|g_j\|_2/2 \text{ for all } g_j\in 2\mathcal F_j\ with\ \|g_j\|_2\geq \delta_{m,j}\}.
		\end{equation*}
	For a $\phi$-mixing process $(X_t)_{t=0}^T$ with $0.781 \leq r_\phi\geq 2$ and $m=T^{\frac{r_\phi}{r_\phi+2}}$, we have 
	$\mathbb P(\mathcal D_{m,T})\geq1-c_2\exp(-c_3(m(\min_j\delta_{m,j}^2))^2)$ where $c_2$, $c_3$ and $c_4$ are constants.
\end{lemma}
Following the outline of the proof for Theorem~\ref{thm:key}, we replace Theorem~\ref{thm:univariate} and Lemma~\ref{lemma:multibound} by Lemma~\ref{lemma:newunivariate} and Lemma~\ref{lemma:newmultibound}, which allows us to prove Theorem~\ref{thm:follow}.\\
\subsection{Proofs for Subsection~\ref{section:twoexamples}}

Now we give proofs for Lemmas~\ref{example:1} and Lemma~\ref{example:2} for the two examples in Subsection~\ref{section:twoexamples}.
\begin{proof}[\bf Proof for Lemma~\ref{example:1}]:
Recall our definition of $\tilde{\epsilon}_m$, by choosing $M_0$ as $M_0=\xi$, we have that
\begin{equation}\begin{aligned}\label{eq:epsilon1}
&\log(dT)\left(3\frac{\log(M_0dT)}{\sqrt{m}}\sqrt{\sum_{i=1}^{M_0}\min(\mu_i,\sigma^2)}+\sqrt{\frac{T}{m}}\sqrt{\sum_{i=M_0+1}^{\infty}\min(\mu_i,\sigma^2)}\right)\\
&=3\frac{\log(\xi dT)\log(dT)}{\sqrt{m}}\sqrt{\sum_{i=1}^{\xi}\min(\mu_i,\sigma^2)}.
\end{aligned}\end{equation}
Since $\min(\mu_i,\sigma^2) \leq \sigma^2$, that equation is upper bounded by
\begin{equation}\label{epsilonub1}
3\sigma\sqrt{\frac{\xi}{m}}\log(\xi dT)\log(dT).
\end{equation}
Since $\tilde{\epsilon}_m$ is the minimal value of $\sigma$ such that~\eqref{eq:epsilon1} lower than $\sigma^2$, from the upper bound~\eqref{epsilonub1} we can show that 
\begin{equation*}
\tilde{\epsilon}_m=O\left(\sqrt{\frac{\xi}{m}}\log(\xi dT)^2\right).
\end{equation*}\end{proof}

\begin{proof}[\bf Proof of Lemma~\ref{example:2}:]
Before proving $\tilde{\epsilon}_m$, we recall the discussion of $\epsilon_m$ \cite{raskutti12}. To simplify the discussion, we assume that there exists an integer $\ell_0$ such that $\sigma^2=\frac{1}{\ell^{2\alpha}_0}$. That assumption doesn't affect the rate which we'll get for $\epsilon_m$. Using the definition of $\ell_0$, $\min(\mu_i,\sigma^2) = \sigma^2$ when $i < \ell_0$ and $\min(\mu_i,\sigma^2) = \mu_i$ when $i \geq \ell_0$. Therefore, since $\mu_i = (1/i)^{2\alpha}$, we have
\begin{equation*}
\frac{1}{\sqrt{m}}\sqrt{\sum_{i=1}^{\infty}\min(\mu_i,\sigma^2)}\leq\frac{1}{\sqrt{m}}\sqrt{\ell_0\sigma^2+\frac{1}{2\alpha-1}\frac{1}{\ell^{2\alpha-1}_0}}
=\frac{1}{\sqrt{m}}\sigma^{1-\frac{1}{2\alpha}}\sqrt{\frac{2\alpha}{2\alpha-1}}.
\end{equation*}
Hence $\epsilon_m=O(m^{-\frac{\alpha}{2\alpha+1}})$. For $\tilde{\epsilon}_m$, we still define $\ell_0$ to be $\sigma^2=\frac{1}{\ell^{2\alpha}_0}$. We require the nuisance parameter $M_0\geq \ell_0$, whose value will be assigned later. Again, using the fact that $\min(\mu_i,\sigma^2) = \sigma^2$ when $i < \ell_0$ and $\min(\mu_i,\sigma^2) = \mu_i$ when $i \geq \ell_0$ and $\mu_i = (1/i)^{2\alpha}$, we have
\begin{equation*}\begin{aligned}
&\log(dT)\left(3\frac{\log(M_0dT)}{\sqrt{m}}\sqrt{\sum_{i=1}^{M_0}\min(\mu_i,\sigma^2)}+\sqrt{\frac{T}{m}}\sqrt{\sum_{i=M_0+1}^{\infty}\min(\mu_i,\sigma^2)}\right)\\
&\leq \log(dT)\left(3\frac{\log(M_0dT)}{\sqrt{m}}\sqrt{\sigma^{2-\frac{1}{\alpha}}+\frac{1}{2\alpha-1}\left(\frac{1}{\ell^{2\alpha-1}_0}-\frac{1}{M_0^{2\alpha-1}}\right)}+\sqrt{\frac{T}{m}}\sqrt{\frac{1}{2\alpha-1}\frac{1}{M_0^{2\alpha-1}}}\right)\\
&=\log(dT)\left(3\frac{\log(M_0dT)}{\sqrt{m}}\sqrt{\frac{2\alpha}{2\alpha-1}\sigma^{2-\frac{1}{\alpha}}-\frac{1}{2\alpha-1}\frac{1}{M_0^{2\alpha-1}}}+\sqrt{\frac{T}{m}}\sqrt{\frac{1}{2\alpha-1}\frac{1}{M_0^{2\alpha-1}}}\right)\\
&\leq \log(dT)\left(3\frac{\log(M_0dT)}{\sqrt{m}}\sqrt{\frac{2\alpha}{2\alpha-1}}\sigma^{1-\frac{1}{2\alpha}}+\sqrt{\frac{T}{m}}\sqrt{\frac{1}{2\alpha-1}}\frac{1}{M_0^{\alpha-\frac{1}{2}}}\right).\end{aligned}\end{equation*}
In order to obtain a similar rate as $\epsilon_m$, we set up the value of $M_0$ such that  $\left(\sqrt{\sqrt{\frac{T}{m}}\frac{1}{M_0^{\alpha-\frac{1}{2}}}}\right)^{1+\frac{1}{2\alpha}}=\frac{1}{\sqrt{m}}$. In other words, $M_0=m^{\frac{1}{2\alpha+1}}T^{\frac{1}{2\alpha-1}}$. After plugging in the value of $M_0$, we obtain an upper bound
\begin{equation}\label{epsilon2}
\log(dT)\left(3\sqrt{\frac{2\alpha}{2\alpha-1}}\frac{\log(M_0dT)}{\sqrt{m}}\sigma^{1-\frac{1}{2\alpha}}+\sqrt{\frac{1}{2\alpha-1}}{\left(\frac{1}{\sqrt{m}}\right)}^{\frac{4\alpha}{2\alpha+1}}\right).
\end{equation}
Compare the upper bound~\eqref{epsilon2} with $\sigma^2$, we obtain
\begin{equation*}
\tilde{\epsilon}_m= O(m^{-\frac{\alpha}{2\alpha+1}}(\log(M_0dT)\log(dT))^{\frac{2\alpha}{2\alpha+1}})=O\left(\left(\frac{\log(dT)^2}{\sqrt{m}}\right)^{\frac{2\alpha}{2\alpha+1}}\right).
\end{equation*}
\end{proof}

\subsection{Proof of Theorem~\ref{theorem:Rademacher}}
We consider single univariate function here and use $f$ to refer each $f_{j,k}$. Finally, we'll use union bound to show that the result holds for every $j,k$. Before presenting the proof, we point out that there exists an equivalent class $\mathbb F$, which means that
\begin{equation}
\sup_{f\in \mathbb F} \left| \frac{1}{T}\sum_{t=1}^T f(X_t)w_t \right|\leq\sup_{\| f \|_{\mathcal H}\leq 1, \| f\|_2\leq \sigma} \left| \frac{1}{T}\sum_{t=1}^T f(X_t)w_t \right|\leq \sup_{f\in \sqrt{2}\mathbb F} \left| \frac{1}{T}\sum_{t=1}^T f(X_t)w_t \right|.
\end{equation}
That function class $\mathbb F$ is defined as
\begin{equation*}
\mathbb F=\{f=\sum_{i=1}^{\infty}\beta_i\sqrt{\mu_i}\Phi_i(x)|\sum_{i=1}^{\infty}\eta_i\beta^2_i\leq 1\} \quad \text{where } \eta_i=\left(\min\left(1,\frac{\sigma^2}{\mu_i}\right)\right)^{-1}.
\end{equation*}
The equivalence is because of
\begin{equation*}
\{f|\| f \|_{\mathcal H}\leq 1, \| f\|_2\leq \sigma\}=\{f=\sum_{i=1}^{\infty}\beta_i\sqrt{\mu_i}\Phi_i(x)|\sum_{i=1}^\infty \beta_i^2\leq 1,\sum_{i=1}^\infty \mu_i\beta_i^2\leq \sigma^2\},
\end{equation*}
(1) $\sum_{i=1}^{\infty}\max(1, \frac{\mu_i}{\sigma^2})\beta^2_i\leq 1\Rightarrow \sum_{i=1}^\infty \beta_i^2\leq 1,\sum_{i=1}^\infty \frac{\mu_i}{\sigma^2}\beta_i^2\leq 1$, and (2) $\sum_{i=1}^\infty \beta_i^2\leq 1,\sum_{i=1}^\infty \frac{\mu_i}{\sigma^2}\beta_i^2\leq 1 \Rightarrow \sum_{i=1}^{\infty}\max(1, \frac{\mu_i}{\sigma^2})\beta^2_i\leq \sum_{i=1}^{\infty}(1+ \frac{\mu_i}{\sigma^2})\beta^2_i\leq 2$. Next, we prove the results for $f\in \mathbb F$. Let's define
\begin{equation}
Y_n=\frac{1}{T}\sum_{t=1}^n \Phi_i(X_t)w_t.
\end{equation}
Then we have
\begin{align*}	
Y_{n}-Y_{n-1}=&\frac{1}{T}\Phi_i(X_n)w_n,\\
E[Y_{n}-Y_{n-1}|w_1,...,w_{n-1}]=&\frac{1}{T}E[\Phi_i(X_n)|w_1,...,w_{n-1}]E[w_n]=0.
\end{align*}
It tells us that $\{Y_n\}_{n=1}^T$ is a martingale. Therefore, we are able to use Lemma 4 on Page 20 in~\cite{hall16}.\\
Additionally, given that $\Phi_i(.)$ is bounded by $1$ and Assumption~\ref{asp:boundednoise} for $w_t$, we know that
\begin{align*}
|Y_n-Y_{n-1}|=&\frac{1}{T}\left|\Phi_i(X_n)w_n\right|\leq \frac{\log(dT)}{T}.
\end{align*}
In order to use Lemma 4 in~\cite{hall16}, we bound the so-called term $M^i_n$ and hence the so-called summation term $D_n$ in~\cite{hall16}, which are
\begin{align*}
M^i_n=&\sum_{t=1}^n E\left[(Y_t-Y_{t-1})^i|w_1,...,w_{t-1}\right]\leq n\frac{\log(dT)^i}{T^i}\text{, and}\\
D_n=&\sum_{i\geq 2} \frac{\varrho^i}{i!}M^i_n\leq \sum_{i\geq 2} \frac{\varrho^i}{i!}n\frac{\log(dT)^i}{T^i}=n\left(e^{\varrho \log(dT) /T}-1-\frac{\varrho \log(dT) }{T}\right),
\end{align*}
for any nuisance parameter $\varrho$. That bound on $D_n$ is defined as $\hat{D}_n$. Then using the results from Lemma 4 in~\cite{hall16} that $\max(E[e^{\varrho Y_n}], E[e^{-\varrho Y_n}])\leq e^{\hat{D}_n}$ for a martingale $\{Y_n\}_{n=1}^T$ and the Markov inequality, we are able to get an upper bound on the desired quantity $Y_n$, that is,
	\begin{equation}\begin{aligned}
	P(|Y_n|\geq y)\leq & E[e^{\varrho |Y_n|}]e^{-\varrho y}\leq \left(E[e^{\varrho Y_n}]+E[e^{-\varrho Y_n}]\right)e^{-\varrho y}\leq 2e^{\hat{D}_n-\varrho y}\\
	= & 2\exp(n(e^{\varrho \log(dT)/T}-1-\frac{\varrho \log(dT)}{T})-\varrho y).
	\end{aligned}\end{equation}
	By setting the nuisance parameter $\varrho=\frac{T}{\log(dT)}\log(\frac{yT}{n\log(dT)}+1)$, that yields the lowest bound
	\begin{equation*}
	P(|Y_n|\geq y)\leq 2\exp\left(-nH\left(\frac{Ty}{n\log(dT)}\right)\right),\; \mbox{where}\; H(x)=(1+x)\log(1+x)-x.
	\end{equation*}
	We can use the fact that $H(x)\geq \frac{3x^2}{2(x+3)}$ for $x\geq 0$ to further simplify the bound and get
	\begin{equation*}
	P(|Y_n|\geq y)\leq 2\exp\left(\frac{-3T^2 y^2}{2Ty\log(dT)+6n\log(dT)^2}\right).
	\end{equation*}
	Plugging in the definition of $Y_n$, this result means that
	\begin{equation*}
	P(\left|\frac{1}{T}\sum_{t=1}^n \Phi_i(X_t)w_t\right|\geq y) \leq 2\exp\left(\frac{-3T^2 y^2}{2Ty\log(dT)+6n\log(dT)^2}\right).
	\end{equation*}
	Then by setting $n=T$, we get
	\begin{equation}
	P(\left|\frac{1}{T}\sum_{t=1}^T \Phi_i(X_t)w_t\right|\geq y) \leq 2\exp\left(\frac{-3T y^2}{2y\log(dT)+6\log(dT)^2}\right).
	\end{equation}
	Using union bound, we obtain an upper bound for the supreme over $M_0$ such terms, which is
	\begin{equation}\label{eq:Radunion}
	P(\sup_{i=1,2,..,M_0} \left|\frac{1}{T}\sum_{t=1}^T \Phi_i(X_t)w_t\right|\geq y) \leq \exp\left(\frac{-3T y^2}{2y\log(dT)+6\log(dT)^2}+\log(2M_0)\right).
	\end{equation}
	We will show next that~\eqref{eq:Radunion} enables us to bound $\sup_{f\in F}\left| \frac{1}{T}\sum_{t=1}^T f(X_t)w_t \right|$, which is our goal. First, we decompose it into two parts
	\begin{align*}
	&\sup_{f\in \mathbb F}\left| \frac{1}{T}\sum_{t=1}^T f(X_t)w_t \right|\\
	&=\sup_{\sum_{i=1}^{\infty}\eta_i\beta^2_i\leq 1} \left|\frac{1}{T}\sum_{t=1}^T \left(\sum_{i=1}^{\infty}\beta_i\sqrt{\mu_i}\Phi_i(X_t)\right)w_t\right|\\
	&=\sup_{\sum_{i=1}^{\infty}\eta_i\beta^2_i\leq 1} \left|\frac{1}{T} \sum_{i=1}^{\infty}\beta_i\sqrt{\mu_i}\left(\sum_{t=1}^T\Phi_i(X_t)w_t\right)\right|\\
	&\leq \sup_{\sum_{i=1}^{\infty}\eta_i\beta^2_i\leq 1} \left|\frac{1}{T} \sum_{i=1}^{M_0}\beta_i\sqrt{\mu_i}\left(\sum_{t=1}^T\Phi_i(X_t)w_t\right)\right|+\sup_{\sum_{i=1}^{\infty}\eta_i\beta^2_i\leq 1} \left|\frac{1}{T} \sum_{i=M_0+1}^{\infty}\beta_i\sqrt{\mu_i}\left(\sum_{t=1}^T\Phi_i(X_t)w_t\right)\right|.
	\end{align*}
	The second part can be easily bounded using Assumption~\ref{asp:boundednoise} in following
	\begin{equation*}
	\sup_{\sum_{i=1}^{\infty}\eta_i\beta^2_i\leq 1} \left|\frac{1}{T} \sum_{i=M_0+1}^{\infty}\beta_i\sqrt{\mu_i}\left(\sum_{t=1}^T\Phi_i(X_t)w_t\right)\right|
	\leq \sup_{\sum_{i=1}^{\infty}\eta_i\beta^2_i\leq 1} \sum_{i=M_0+1}^{\infty}\beta_i\sqrt{\mu_i}\log(dT).
	\end{equation*}
	Using Cauchy-Schwarz inequality, this upper bound is further bounded by
	\begin{equation*}
	\sup_{\sum_{i=1}^{\infty}\eta_i\beta^2_i\leq 1} \sqrt{\sum_{i=M_0+1}^{\infty}\eta_i\beta^2_i}\sqrt{\sum_{i=M_0+1}^{\infty}\frac{\mu_i}{\eta_i}}\log(dT),
	\end{equation*}
	which is smaller than
	\begin{equation*}
	\sqrt{\sum_{i=M_0+1}^{\infty}\frac{\mu_i}{\eta_i}}\log(dT)=\sqrt{\sum_{i=M_0+1}^{\infty}\min(\mu_i,\sigma^2)}\log(dT).
	\end{equation*}
	Our next goal is to show that we can bound the first part $\sup_{\sum_{i=1}^{\infty}\eta_i\beta^2_i\leq 1} \left|\frac{1}{T} \sum_{i=1}^{M_0}\beta_i\sqrt{\mu_i}\left(\sum_{t=1}^T\Phi_i(X_t)w_t\right)\right|$ using~\eqref{eq:Radunion}. To bound that, simply using Cauchy-Schwarz inequality, we get
	\begin{align*}
	&\sup_{\sum_{i=1}^{\infty}\eta_i\beta^2_i\leq 1} \left|\frac{1}{T} \sum_{i=1}^{M_0}\beta_i\sqrt{\mu_i}\left(\sum_{t=1}^T\Phi_i(X_t)w_t\right)\right|\\
	&\leq \sup_{\sum_{i=1}^{\infty}\eta_i\beta^2_i\leq 1} \sqrt{\sum_{i=1}^{M_0}\eta_i\beta^2_i}\sqrt{\sum_{i=1}^{M_0} \frac{\mu_i}{\eta_i}\left(\frac{1}{T}\sum_{t=1}^T\Phi_i(X_t)w_t\right)^2}\\
	&\leq \sqrt{\sum_{i=1}^{M_0} \frac{\mu_i}{\eta_i}}\sup_{i=1,2,..,M_0} \left|\frac{1}{T}\sum_{t=1}^T \Phi_i(X_t)w_t\right|.
	\end{align*}
	Using \eqref{eq:Radunion}, we show that the first part is upper bounded by 
	\begin{equation*}
	\sqrt{\sum_{i=1}^{M_0}\min(\mu_i,\sigma^2)}y,
	\end{equation*}
	with probability at least $1-\exp\left(\frac{-3T y^2}{2y\log(dT)+6\log(dT)^2}+\log(M_0)\right)$.
	Therefore, after combining the bounds on the two parts, we obtain the upper bound for $\sup_{f\in \sqrt{2}\mathbb F} \left| \frac{1}{T}\sum_{t=1}^T f(X_t)w_t \right|$, which is
	\begin{equation*}
	\sup_{f\in \sqrt{2}\mathbb F} \left| \frac{1}{T}\sum_{t=1}^T f(X_t)w_t \right|\leq \sqrt{2}\left(\sqrt{\sum_{i=1}^{M_0}\min(\mu_i,\sigma^2)}y+\sqrt{\sum_{i=M_0+1}^{\infty}\min(\mu_i,\sigma^2)}\log(dT)\right),
	\end{equation*}
	with probability at least $1-\exp\left(\frac{-3T y^2}{2y\log(dT)+6\log(dT)^2}+\log(M_0)\right)$.
	Further, after applying union bound on all $(j,k)\in \{1,2,...,d\}^2$ and recalling the connection between $\{f|\|f\|_{\mathcal H}\leq 1,\|f\|_2\leq\sigma\}$ and $\sqrt{2}\mathbb F$, we can show that with probability at least $1-\exp\left(\frac{-3T y^2}{2y\log(dT)+6\log(dT)^2}+\log(M_0)+2\log(d)\right)$,
	\begin{equation*}
	\sup_{\| f_{j,k} \|_{\mathcal H}\leq 1 \| f_{j,k}\|_2\leq \sigma} \left| \frac{1}{T}\sum_{t=1}^T f_{j,k}(X_t)w_{t,j} \right|\leq \sqrt{2}\left(\sqrt{\sum_{i=1}^{M_0}\min(\mu_i,\sigma^2)}y+\sqrt{\sum_{i=M_0+1}^{\infty}\min(\mu_i,\sigma^2)}\log(dT)\right),
	\end{equation*}
	for all $(j,k)\in\{1,2,...,d\}^2$ and any $M_0,y$.
	

	Finally, by setting $y=3\frac{(\log(M_0dT))\log(dT)}{\sqrt{T}}$, we obtain that, with probability at least $1-\frac{1}{M_0T}$,
	\begin{equation*}\begin{aligned}
	&\sup_{\| f_{j,k} \|_{\mathcal H}\leq 1, \| f_{j,k}\|_2\leq \sigma} \left| \frac{1}{T}\sum_{t=1}^T f_{j,k}(X_t)w_t \right|\\
	&\leq \sqrt{2}\log(dT)\left(3\frac{(\log(M_0dT))}{\sqrt{T}}\sqrt{\sum_{i=1}^{M_0}\min(\mu_i,\sigma^2)}+\sqrt{\sum_{i=M_0+1}^{\infty}\min(\mu_i,\sigma^2)}\right).
	\end{aligned}\end{equation*}
	Here, we assumed $T\geq 2$, $\log(M_0dT)\geq 1$. Our definition of $\tilde{\epsilon}_m$ guarantees that, if $\sigma>\tilde{\epsilon}_m$, then
	\begin{equation*} \sqrt{2}\log(dT)\left(3\frac{(\log(M_0dT))}{\sqrt{T}}\sqrt{\sum_{i=1}^{M_0}\min(\mu_i,\sigma^2)}+\sqrt{\sum_{i=M_0+1}^{\infty}\min(\mu_i,\sigma^2)}\right)\leq \sqrt{2}\sqrt{\frac{m}{T}}\sigma^2.
	\end{equation*}
	That completes our proof for Theorem~\ref{theorem:Rademacher}.
\subsection{Proof of Theorem~\ref{thm:univariate}}
	Since we have $\|f_{j,k}\|_{\infty}\leq 1$, it suffices to bound
	\begin{equation*}
	 \mathbb P\left(\sup_{j,k}\sup_{f_{j,k}\in B_{\mathcal H}(1), \|f_{j,k}\|_2\leq \gamma_m} |\|f_{j,k}\|^2_T-\|f_{j,k}\|^2_2|\geq \frac{\gamma^2_m}{4}\right).
	 \end{equation*}
	The proofs are based on the result for independent case from Lemma 7 in \cite{raskutti12}, which shows that there exists constants $(\tilde{c}_1,\tilde{c}_2)$ such that
	\begin{equation}\label{thm4ind}
	\mathbb P_0\left(\sup_{j,k}\sup_{f_{j,k}\in B_H(1), \|f_{j,k}\|_2\leq \gamma_m} |\frac{1}{m}\sum_{t=1}^m f_{j,k}^2(\tilde{X}_t)-\|f_{j,k}\|^2_2|\geq \frac{\gamma^2_m}{10}\right)\leq \tilde{c}_1\exp(-\tilde{c}_2m\gamma_m^2),
	\end{equation}
	where $\{\tilde{X}_t\}_{t=1}^m$ are i.i.d drawn from the stationary distribution of $X_t$ denoted by $\mathbb P_0$. Let $T=m\ell$. We divide the stationary $T$-sequence $X_T=(X_1,X_2,...,X_T)$ into $m$ blocks of length $\ell$. We use $X_{a,b}$ to refer the $b$-th variable in block $a$. Therefore, we can rewrite 
	\begin{equation*}
	\mathbb P\left(\sup_{j,k}\sup_{f_{j,k}\in B_{\mathcal H}(1), \|f_{j,k}\|_2\leq \gamma_m} |\frac{1}{T}\sum_{t=1}^T f_{j,k}^2(X_t)-\|f_{j,k}\|^2_2|\geq \frac{\gamma^2_m}{4}\right)   
	\end{equation*}
	as
	\begin{equation}\label{eqthm4}
	\mathbb P\left(\sup_{j,k}\sup_{f_{j,k}\in B_{\mathcal H}(1), \|f_{j,k}\|_2\leq \gamma_m} |\frac{1}{\ell}\sum_{b=1}^\ell\frac{1}{m}\sum_{a=1}^m f_{j,k}^2(X_{a,b})-\|f_{j,k}\|^2_2|\geq \frac{\gamma^2_m}{4}\right).
	\end{equation}
	Using the fact that $\sup|\sum..|\leq \sum\sup|..|$, \eqref{eqthm4} is smaller than
	\begin{equation*}
	\mathbb P\left(\frac{1}{\ell}\sum_{b=1}^\ell\sup_{j,k}\sup_{f_{j,k}\in B_{\mathcal H}(1), \|f_{j,k}\|_2\leq \gamma_m} |\frac{1}{m}\sum_{a=1}^m f_{j,k}^2(X_{a,b})-\|f_{j,k}\|^2_2|\geq \frac{\gamma^2_m}{4}\right),
	\end{equation*}
	which, by using the fact that $P(\frac{1}{\ell}\sum_{i=1}^\ell a_i\geq c)\leq P(\cup_{i=1}^\ell (a_i\geq c))\leq \sum_{i=1}^\ell P(a_i\geq c)$, is bounded by
	\begin{equation*}
    \sum_{b=1}^\ell \mathbb P\left(\sup_{j,k}\sup_{f_{j,k}\in B_{\mathcal H}(1), \|f_{j,k}\|_2\leq \gamma_m} |\frac{1}{m}\sum_{a=1}^m f_{j,k}^2(X_{a,b})-\|f_{j,k}\|^2_2|\geq \frac{\gamma^2_m}{4}\right).
	\end{equation*}
	Using the fact that the process is stationary, it is equivalent to
	\begin{equation}\label{eqthm4sec}
	\ell \mathbb P\left(\sup_{j,k}\sup_{f_{j,k}\in B_H(1), \|f_{j,k}\|_2\leq \gamma_m} |\frac{1}{m}\sum_{a=1}^m f_{j,k}^2(X_{a,\ell})-\|f_{j,k}\|^2_2|\geq \frac{\gamma^2_m}{4}\right).
	\end{equation}
	Our next steps are trying to bound the non-trivial part in~\eqref{eqthm4sec}.
	Because of Lemma 2 in \cite{nobel93}, we can replace $\{X_{a,l}\}_{a=1}^m$ by their independent copies under probability measure $P_0$ with a sacrifice of $m\beta(\ell)$. Then we are able to use~\eqref{thm4ind} to bound the remaining probability. First, using Lemma 2 in \cite{nobel93}, we have
	\begin{align*}
	&\mathbb P\left(\sup_{j,k}\sup_{f_{j,k}\in B_H(1), \|f_{j,k}\|_2\leq \gamma_m} |\frac{1}{m}\sum_{a=1}^m f_{j,k}^2(X_{a,\ell})-\|f_{j,k}\|^2_2|\geq \frac{\gamma^2_m}{4}\right)\\ 
	&\leq \mathbb P_0\left(\sup_{j,k}\sup_{f_{j,k}\in B_{\mathcal H}(1), \|f_{j,k}\|_2\leq \gamma_m} |\frac{1}{m}\sum_{a=1}^m f_{j,k}^2(X_{a,\ell})-\|f_{j,k}\|^2_2|\geq \frac{\gamma^2_m}{4}\right)+m\beta(\ell).
	\end{align*}
	Now, using~\eqref{thm4ind}, it is bounded by
	\begin{equation*}
	\tilde{c}_1\exp(2\log (d)-\tilde{c}_2m\gamma^2_m)+m\beta(\ell).
	\end{equation*}
	Therefore, we get
	\begin{align*}
	&\mathbb P\left(\sup_{j,k}\sup_{f_{j,k}\in B_{\mathcal H}(1), \|f_{j,k}\|_2\leq \gamma_m} |\frac{1}{T}\sum_{t=1}^T f_{j,k}^2(X_t)-\|f_{j,k}\|^2_2|\geq \frac{\gamma^2_m}{4}\right)\\
	&\leq \ell \mathbb P\left(\sup_{j,k}\sup_{f_{j,k}\in B_{\mathcal H}(1), \|f_{j,k}\|_2\leq \gamma_m} |\frac{1}{m}\sum_{a=1}^m f_{j,k}^2(X_{a,\ell})-\|f_{j,k}\|^2_2|\geq \frac{\gamma^2_m}{4}\right)\\
	&\leq \ell \tilde{c}_1\exp(2\log (d)-\tilde{c}_2m\gamma^2_m)+T\beta(\ell).
	\end{align*}
	Recall that $\ell = T/m$ and the definition of $\beta(\ell)$, which is equal to $\ell^{-r_\beta}$, that bound hence is
	\begin{equation*}
	\tilde{c}_1\exp(2\log(dT)-\tilde{c}_2m\gamma_m^2)+T\left(\frac{T}{m}\right)^{-r_\beta}.
	\end{equation*}
	Recall our definition of $\gamma_m$ with $m\gamma_m^2\geq c_1^2\log(dT)$ and $m=T^{\frac{c_0r_\beta-1}{c_0r_\beta}}$, hence that probability is $$\left(\tilde{c}_3\exp(-\tilde{c}_4m\gamma_m^2)+T^{-\left(\frac{1-c_0}{c_0}\right)}\right),$$ for some constants $\tilde{c}_3$ and $\tilde{c}_4$. That completes the proof. For the follow-up statement, condition on the event $\mathcal B_{m,T}$, for any $g\in B_{\mathcal H}(1)$ with $\|g\|_2\geq \gamma_m$, we have $h=\gamma_m\frac{g}{\|g\|_2}$ is in $B_{\mathcal H}(1)$ and $\|h\|_2\leq \gamma_m$. Therefore, we have 
	$$\left|\|\gamma_m\frac{g}{\|g\|_2}\|_T-\|\gamma_m\frac{g}{\|g\|_2}\|_2\right|\leq \frac{\gamma_m}{2},$$
	which implies
	$$|\|g\|_T-\|g\|_2|\leq \frac{1}{2}\|g\|_2.$$

\subsection{Other proofs}\label{otherproofs}
\begin{proof}[{\bf Proof of Lemma \ref{lemma:unibound}}]
	The statement which we want to show is equivalent to 
	\begin{equation}\label{eq:unigoal}
	|\frac{1}{T}\sum_{t=1}^Tf_{j,k}(X_t)w_{t,j}|\leq 4\sqrt{2}\| f_{j,k}\|_{\mathcal H}\sqrt{\frac{m}{T}}(\tilde\gamma_m^2 + \tilde\gamma_m\frac{\| f_{j,k}\|_T}{\| f_{j,k}\|_{\mathcal H}} ) 
	\end{equation} 
	for any $f_{j,k}\in \mathcal H$, for any $(j,k)\in[1,2,...,d]^2$.\\
	For each $j,k$, we define
	\begin{equation*}
	Z_{T,j,k}(w;\ell) := |\sup_{\| f_{j,k} \|_T\leq \ell, \| f_{j,k} \|_{\mathcal H}\leq 1}\frac{1}{T}\sum_{t=1}^T f_{j,k}(X_t)w_{t,j}|.
	\end{equation*}
	We claim that on event $\mathcal A_{m,T}\cap \mathcal B_{m,T}$,
	\begin{equation}\label{eq:boundZT}
	Z_{T,j,k}(w;\ell)\leq 4\sqrt{2}\sqrt{\frac{m}{T}}(\tilde\gamma^2_m+\tilde\gamma_m \ell)\text{ for any }(j,k)\in[1,2,...,d]^2.
	\end{equation}
	We give the proof in following.
	\begin{proof}
	Based on the sandwich inequality in Theorem \ref{thm:univariate}, for any $g\in B_{\mathcal H}(1)$, any $\sigma\geq \gamma_m$, when $\|g\|_2\geq 2\sigma \geq \gamma_m$, $\|g\|_T\geq \frac{\|g\|_2}{2}\geq \sigma$. Therefore, 
	\begin{equation}\label{lemma3key}
	\text{For any }\sigma\geq \gamma_m,\text{ if }\|g\|_T\leq \sigma\text{ then }\|g\|_2\leq 2\sigma. 
	\end{equation}
	Using this fact, we proceed the proof in two cases.\\ 
	{\bf Case 1:} If $\ell \leq \tilde\gamma_m$, then 
	\begin{align*}
	&Z_{T,j,k}(w;\ell) = |\sup_{\| f_{j,k} \|_T\leq \ell, \| f_{j,k} \|_{\mathcal H}\leq 1}\frac{1}{T}\sum_{t=1}^T f_{j,k}(X_t)w_{t,j}|\\
	&\leq |\sup_{\|f_{j,k}\|_T\leq \tilde\gamma_m, \|f_{j,k}\|_{\mathcal H}\leq 1}\frac{1}{T}\sum_{t=1}^T f_{j,k}(X_t)w_{t,j}|.
	\end{align*}
	Since $\tilde{\gamma}_m\geq \gamma_m$, using the fact~\eqref{lemma3key}, we get
	\begin{equation*}
	Z_{T,j,k}(w;\ell)\leq |\sup_{\|f_{j,k}\|_2\leq 2\tilde\gamma_m, \|f_{j,k}\|_{\mathcal H}\leq 1}\frac{1}{T}\sum_{t=1}^T f_{j,k}(X_t)w_{t,j}|.
	\end{equation*}
	Further, since $\tilde{\gamma}_m\geq \tilde{\epsilon}_m$, we are able to use Theorem~\ref{theorem:Rademacher} and show that
	\begin{equation*}
	Z_{T,j,k}(w;\ell)\leq 4\sqrt{2}\sqrt{\frac{m}{T}}\tilde\gamma_m^2.
	\end{equation*}

	{\bf Case 2:} If $\ell\geq \tilde\gamma_m$, we use scaling on $f$ to transform it to Case 1, hence we can show a bound in following.
	\begin{align*}
	&Z_{T,j,k}(w;\ell) = |\sup_{\| f_{j,k} \|_T\leq \ell, \| f_{j,k} \|_{\mathcal H}\leq 1}\frac{1}{T}\sum_{t=1}^T f_{j,k}(X_t)w_{t,j}| \\
	&= |\frac{\ell}{\tilde\gamma_m}\sup_{\|\frac{\tilde\gamma_m}{\ell} f_{j,k} \|_T\leq \tilde\gamma_m, \|\frac{\tilde\gamma_m}{\ell} f_{j,k} \|_{\mathcal H}\leq \frac{\tilde\gamma_m}{\ell}}\frac{1}{T}\sum_{t=1}^T \frac{\tilde\gamma_m}{\ell}f_{j,k}(X_t)w_{t,j}|\\
	&\leq |\frac{\ell}{\tilde\gamma_m}\sup_{\| \tilde{f}_{j,k} \|_T\leq \tilde\gamma_m, \| \tilde{f}_{j,k} \|_{\mathcal H}\leq 1} \frac{1}{T}\sum_{t=1}^T \tilde{f}_{j,k}(X_t)w_{t,j}|\\
	&\leq 4\sqrt{2}\sqrt{\frac{m}{T}}\ell\tilde\gamma_m.
	\end{align*}

	Therefore, statement \eqref{eq:boundZT} is true.
	\end{proof}
	Next, we use proof by contradiction to prove~\eqref{eq:unigoal}. If \eqref{eq:unigoal} fails for a function $f^0_{j,k}$, we can assume $\|f^0_{j,k}\|_{\mathcal H}=1$, otherwise, statement also fails for $\frac{f^0_{j,k}}{\|f^0_{j,k}\|_{\mathcal H}}$. Then we let $\ell=\|f^0_{j,k}\|_T$. Now $\|f^0_{j,k}\|_T\leq \ell$, $\|f^0_{j,k}\|_{\mathcal H}\leq 1$, but
	\begin{equation*}
	\left|\frac{1}{T}\sum_{t=1}^Tf^0_{j,k}(X_t)w_{t,j}\right|\geq 4\sqrt{2}\sqrt{\frac{m}{T}}(\tilde\gamma_m^2\|f^0_{j,k}\|_H+\tilde\gamma_m\|f^0_{j,k}\|_T)=4\sqrt{2}\sqrt{\frac{m}{T}}(\tilde\gamma_m^2+\tilde\gamma_m \ell),
	\end{equation*}
	which contradicts \eqref{eq:boundZT}. Therefore, \eqref{eq:unigoal} is true.
\end{proof}
\begin{proof}[{\bf Proof of Lemma \ref{lemma:unidifference}}]
	First, using Theorem~\ref{thm:univariate}, on event $\mathcal B_{m,T}$ for any $(j,k)\in[1,2,...,d]^2$,
	\begin{equation}
	\|f_{j,k}\|_T\leq \|f_{j,k}\|_2+\frac{\gamma_m}{2} \text{ for all } f_{j,k}\in B_{\mathcal H}(1)\ and\ \|f_{j,k}\|_2\leq \gamma_m.
	\end{equation}
	On the other hand, if $\|f_{j,k}\|_2>\gamma_m$, then the sandwich relation in Theorem~\ref{thm:univariate} implies that $\|f_{j,k}\|_T\leq 2\|f_{j,k}\|_2$. Therefore, we have
	\begin{equation*}
	\|f_{j,k}\|_T\leq 2\|f_{j,k}\|_2+\frac{\gamma_m}{2} \text{ for all } f_{j,k}\in B_{\mathcal H}(1).
	\end{equation*}
	The proof is completed by noticing $g_{j,k}=2f_{j,k}$.
\end{proof}
\begin{proof}[{\bf Proof of Lemma \ref{lemma:multibound}}]
	First, we point out that we only need to show that
	$$\|g_j\|_T\geq \delta_{m,j}/2 \text{ for all } g_j\in 2\mathcal F_j\ with\ \|g_j\|_2=\delta_{m,j},$$
	because if $\|g_j\|_2 \geq \delta_{m,j}$, we can scale $g_j$ to $\frac{\delta_{m,j}}{\|g_j\|_2}g_j$, which belongs to $2\mathcal F_j$ as well since $\frac{\delta_{m,j}}{\|g_j\|_2}< 1$.
	We choose a truncation level $\tau>0$ and define the function
	$$\ell_\tau(u)=\{\begin{array}{cc}
	u^2 & if\ |u|\leq \tau\\
	\tau^2 & otherwise
	\end{array}$$
	Since $u^2\geq \ell_\tau(u)$ for all $u\in \mathbb R$, we have
	$$\frac{1}{T}\sum_{t=1}^Tg_j^2(X_t)\geq \frac{1}{T}\sum_{t=1}^T\ell_\tau(g_j(X_t)).$$
	The remainder of the proof consists of the following steps:

	(1) First, we show that for all $g_j\in 2\mathcal F$ with $\|g_j\|=\delta_{m,j}$, we have
	$$\mathbb E[\ell_\tau(g_j(x))]\geq \frac{1}{2}\mathbb E[g_j^2(x)]=\frac{\delta_{m,j}^2}{2}.$$
	
	(2) Next we prove that
	\begin{equation}
	\sup_{g_j\in 2\mathcal F_j, \|g_j\|_2\leq \delta_{m,j}}|\frac{1}{T}\sum_{t=1}^T\ell_\tau(g_j(X_t))-\mathbb E[\ell_\tau(g(X_t))]|\leq \frac{\delta_{m,j}^2}{4},
	\end{equation}
	with high probability for $\beta$ mixing process with $r\geq 1/c_0$. 
	
	Putting together the pieces, we conclude that for any $g_j\in \mathcal F_j$ with $\|g_j\|_2=\delta_{m,j}$, we have
	$$\frac{1}{T}\sum_{t=1}^Tg_j^2(X_t)\geq \frac{1}{T}\sum_{t=1}^T\ell_\tau(g_j(X_t))\geq \frac{\delta_{m,j}^2}{2}-\frac{\delta_{m,j}^2}{4}=\frac{\delta_{m,j}^2}{4},$$
	with high probability (to be specified later). This shows that event $\mathcal D_{m,T}$ holds with high probability, thereby completing the proof. It remains to establish the claims.\\
	{\bf Part 1. Establishing the lower bound for $\mathbb E[\ell_\tau(g_j(x))]$:}
	\begin{proof}
	 We can not use the same proofs as in the independent case from~\cite{raskutti12}, since each element from the multivariate variable $x=(x_1,...,x_d)$ is not independent from others in the stationary distribution. That is the reason why we need to have Assumption~\ref{asp:fourth}. In the independent case, Assumption~\ref{asp:fourth} is shown to be true in~\cite{raskutti12}. Note that
	 $$g_j(x)=\sum_{k\in U}g_{j,k}(x_j),$$
	 for a subset $U$ of cardinality at most $2s_j$, we have
	 \begin{equation*}
	 \mathbb E[\ell_\tau(g_j(x))]\geq
	 E[g_j^2(x)I[|g_j(x)|\leq \tau]]=
	 \delta_{m,j}^2-E[g_j^2(x)I[|g_j(x)|\geq \tau]].
	 \end{equation*}
	 Using Cauchy-Schwarz inequality and Markov inequality, we can show that
	 \begin{equation*}
	 (\mathbb E[g_j^2(x)I[|g_j(x)|\geq \tau]])^2\leq \mathbb E[g_j^4(x)]P(|g_j(x)|\geq \tau)\leq \mathbb E[g_j^4(x)]\frac{\delta_{m,j}^2}{\tau^2}.
	 \end{equation*}
	 Since $E[g_j^4(x)]\leq C\delta_{m,j}^2=CE[g_j^2(x)]$ given by Assumption~\ref{asp:fourth}, by choosing $\tau\geq 2\sqrt{C}$, we are able to show that $$\mathbb E[\ell_\tau(g_j(x))]\geq \frac{1}{2}\mathbb E[g_j^2(x)]=\frac{\delta_{m,j}^2}{2}.$$
	 \end{proof}
	 
	 {\bf Part 2. Establishing the  probability bound on}
	 \begin{equation*}
	 \sup_{g_j\in 2\mathcal F_j, \|g_j\|_2\leq \delta_{m,j}}|\frac{1}{T}\sum_{t=1}^T\ell_\tau(g_j(X_t))-\mathbb E[\ell_\tau(g_j(X_1))]|\leq \frac{\delta_{m,j}^2}{4}.
	 \end{equation*}
	 \begin{proof}
	 Similar as the proof of Lemma. \ref{lemma:unibound}, we base our proof on the independent result from Lemma 4 in \cite{raskutti12}, which is
	 \begin{equation}\label{lemma5ind}
	 \mathbb P_0(\sup_{g_j\in 2\mathcal F_j, \|g_j\|_2\leq \delta_{m,j}}|\frac{1}{m}\sum_{a=1}^m\ell_\tau(g(X_a))-E[\ell_\tau(g(X_1))]|\geq \frac{\delta_{m,j}^2}{12})\leq \tilde{c}_1\exp(-\tilde{c}_2m\delta_{m,j}^2).
	 \end{equation}
	 We let $T=m\ell$. Using the same facts and results as in the proof for Theorem~\ref{thm:univariate}, we have
	 \begin{align*}
	 &\mathbb P(\sup_{g_j\in 2\mathcal F_j, \|g_j\|_2\leq \delta_{m,j}}|\frac{1}{T}\sum_{t=1}^T\ell_\tau(g_j(X_t))-\mathbb E[\ell_\tau(g_j(X_1))]|\geq \frac{\delta_{m,j}^2}{4})\\
	 &=\mathbb P(\sup_{g_j\in 2\mathcal F_j, \|g_j\|_2\leq \delta_{m,j}}|\frac{1}{\ell}\sum_{b=1}^\ell\frac{1}{m}\sum_{a=1}^m\ell_\tau(g_j(X_{a,b}))-\mathbb E[\ell_\tau(g(X_{a,1}))]|\geq \frac{\delta_{m,j}^2}{4})\\
	 &\leq \ell P_0(\sup_{g_j\in 2\mathcal F_j, \|g_j\|_2\leq \delta_{m,j}}|\frac{1}{m}\sum_{a=1}^m\ell_\tau(g_j(X_{a,\ell}))-\mathbb E[\ell_\tau(g(X_{a,\ell}))]|\geq \frac{\delta_{m,j}^2}{4})+T\beta(\ell).
	 \end{align*}
	 Using~\eqref{lemma5ind}, we conclude that it is upper bounded by
	 \begin{equation*}
	 \ell \tilde{c}_1\exp{(-\tilde{c}_2m\delta_{m,j}^2)}+T(\frac{T}{m})^{-r_\beta},
	 \end{equation*}
	 which is then upper bounded by $\tilde{c}_3\exp(-\tilde{c}_4m\delta_{m,j}^2)+T^{-\frac{1-c_0}{c_0}}$ for constants $\tilde{c}_3, \tilde{c}_4$.
	 \end{proof}
	 Now, we proved that all claims are correct. Therefore, we complete the proof.
\end{proof}
\begin{proof}[{\bf Proof of Lemma \ref{lemma:newunivariate}}]
    For $\phi$-mixing process with $0.781\leq \phi\leq 2$, we can use the concentration inequality from~\cite{kontorovich08} to show sharper rate in Lemma~\ref{lemma:newunivariate} than Theorem~\ref{thm:univariate}. That concentration inequality is presented in following.
    \begin{lemma}[McDirmaid inequality in \cite{kontorovich08, mohri10}]\label{McD}
		Suppose $\mathbb S$ is a countable space, $\mathbb F_{\mathbb S}$ is the set of all subsets of $\mathbb S^n$, $\mathbb Q$ is a probability measure on $(\mathbb S^n,\mathbb F_{\mathbb S})$ and $g:\mathbb S^n\rightarrow \mathbb R$ is a $c$-Lipschitz function (with respect to the Hamming metric) on $\mathbb S^n$ for some $c>0$. Then for any $y>0,$
		$$\mathbb P(|g(X)-\mathbb Eg(X)|\geq y)\leq 2\exp\left(-\frac{y^2}{2nc^2(1+2\sum_{\ell=1}^{n-1}\phi(\ell))^2}\right).$$
	\end{lemma}
	Its original version is for discrete space, which is then generalized to continuous case in \cite{kontorovich07}. Here, we use its special form for the $\phi$-mixing process which is pointed out in \cite{kontorovich07} and \cite{mohri10}.
	
	For our statement, as pointed out in the proof for Theorem~\ref{thm:univariate}, since we have $\|f_{j,k}\|_{\infty}\leq 1$, it suffices to bound
	\begin{equation}\label{lemma6ind}
	 \mathbb P\left(\sup_{j,k}\sup_{f_{j,k}\in B_{\mathcal H}(1), \|f_{j,k}\|_2\leq \gamma_m} |\|f_{j,k}\|^2_T-\|f_{j,k}\|^2_2|\geq \frac{\gamma^2_m}{4}\right).
	 \end{equation}
	The proofs are based on independent result from Lemma 7 in \cite{raskutti12}, which shows that there exists constants $(\tilde{c}_1,\tilde{c}_2)$ such that
	\begin{equation*}
	\mathbb P_0\left(\sup_{j,k}\sup_{f_{j,k}\in B_H(1), \|f_{j,k}\|_2\leq \gamma_m} |\frac{1}{m}\sum_{t=1}^m f_{j,k}^2(\tilde{X}_t)-\|f_{j,k}\|^2_2|\geq \frac{\gamma^2_m}{10}\right)\leq \tilde{c}_1\exp(-\tilde{c}_2m\gamma_m^2),
	\end{equation*}
	where $\{\tilde{X}_t\}_{t=1}^m$ are i.i.d drawn from the stationary distribution of $X_t$ denoted by $\mathbb P_0$. 
	
	Now, we can use Lemma~\ref{McD} to show the sharper rate. Recall that $\|f_{j,k}\|_{\infty}\leq 1$, we define
	$$g(X)=\sup_{f_{j,k}\in B_{\mathcal H}(1), \|f_{j,k}\|_2\leq \gamma_m} |\|f_{j,k}\|^2_T-\|f_{j,k}\|^2_2|.$$
	Then,
	\begin{equation*}
	|g(X)-g(Y)|\leq \sup_{f_{j,k}\in B_{\mathbb H}(1), \|f_{j,k}\|_2\leq \gamma_m}|\frac{1}{T}\sum_{t=1}^T f_{j,k}(X_t)^2-\frac{1}{T}\sum_{t=1}^T f_{j,k}(Y_t)^2|\leq \frac{1}{T}dist_{\mathcal HM}(X,Y),
	\end{equation*}
	where $dist_{\mathcal HM}(X,Y)$ means the Hamming metric between $X$ and $Y$, which equals to how many paired elements are different between $X$ and $Y$.
	Thus, we know that $g(X)$ is $\frac{1}{T}$-Lipschitz with respect to the Hamming metric. 
	Therefore, using Lemma~\ref{McD}, we show that
	\begin{equation*}
	\mathbb P(|g(X)-\mathbb Eg(X)|\geq \frac{\gamma^2_m}{8})\leq 2\exp(-\frac{T\gamma^4_m}{128(1+2\sum_{\ell=1}^{T-1}\phi(\ell))^2}).
	\end{equation*}
	Using the fact that $\phi(\ell)=\ell^{-r_\phi}$, we show that probability is bounded by   $O(\exp(-\min(T^{2r_\phi-1},T)\gamma^4_m))$. If we use union bound on $d^2$ terms, that is at most $O(\exp(2\log(d)-\min(T^{2r_\phi-1},T)\gamma_m^4))$. Since $0.781\leq r_\phi\leq 2$, we show that $T^{2r_\phi-1}\gamma_m^4 = (m^{\frac{r_\phi+2}{r_\phi}})^{2r_\phi-1}\gamma_m^4=\Omega (m^2\gamma_m^4)$ and  $T\gamma_m^4=(m^{\frac{r_\phi+2}{r_\phi}})\gamma_m^4= \Omega(m^2\gamma_m^4)$. Therefore, the probability is at most $c_2\exp(-c_3(m\gamma_m^2)^2)$ for some constants $(c_2,c_3)$.
	
	The remaining proof is then to show that $\mathbb Eg(X) \leq \frac{\gamma^2_m}{8}$. In other words, we need to show that for sufficient large $m$,
	\begin{equation}
	\mathbb E\sup_{f_{j,k}\in B_{\mathcal H}(1), \|f_{j,k}\|_2\leq \gamma_m} |\frac{1}{T}\sum_{t=1}^T f_{j,k}^2(X_t)-\|f_{j,k}\|^2_2|\leq \frac{\gamma^2_m}{8}
	\end{equation}
	First, we use the same fact and results as in the proof for Theorem~\ref{thm:univariate} to show that
	\begin{align*}
	&\mathbb E[\sup_{f_{j,k}\in B_{\mathcal H}(1), \|f_{j,k}\|_2\leq \gamma_m} |\frac{1}{T}\sum_{t=1}^Tf_{j,k}^2(X_t)-\|f_{j,k}\|^2_2|]\\
	&=\mathbb E[\sup_{f_{j,k}\in B_{\mathcal H}(1), \|f_{j,k}\|_2\leq \gamma_m} |\frac{1}{\ell}\sum_{b=1}^\ell\frac{1}{m}\sum_{a=1}^m f_{j,k}^2(X_{a,b})-\|f_{j,k}\|^2_2|]\\
	&\leq \frac{1}{\ell}\sum_{b=1}^\ell \mathbb E[\sup_{f_{j,k}\in B_{\mathcal H}(1), \|f_{j,k}\|_2\leq \gamma_m} |\frac{1}{m}\sum_{a=1}^m f_{j,k}^2(X_{a,b})-\|f_{j,k}\|^2_2|]\\
	&= \mathbb E[\sup_{f_{j,k}\in B_{\mathcal H}(1), \|f_{j,k}\|_2\leq \gamma_m} |\frac{1}{m}\sum_{a=1}^m f_{j,k}^2(X_{a,\ell})-\|f_{j,k}\|^2_2|].
	\end{align*}
	Using the fact that $\mathbb E[Z] = \mathbb E[ZI(Z\leq \delta)] + \mathbb E[ZI(Z\geq \delta)]\leq \delta + \|Z\|_{\infty}\mathbb P(Z\geq \delta)$ and $\|f_{j,k}\|_{\infty}\leq 1$, we show an upper bound
	\begin{equation*}
	\delta+2\mathbb P(\sup_{f_{j,k}\in B_{\mathcal H}(1), \|f_{j,k}\|_2\leq \gamma_m} |\frac{1}{m}\sum_{a=1}^m f_{j,k}^2(X_{a,\ell})-\|f_{j,k}\|^2_2|\geq \delta)\text{ for any }\delta>0.
	\end{equation*}
	As in the proof of Theorem~\ref{thm:univariate}, we use Lemma 2 in \cite{nobel93} to connect the dependence probability with independence probability, which gives us 
	\begin{align*}
	&\delta+2\mathbb P(\sup_{f_{j,k}\in B_{\mathcal H}(1), \|f_{j,k}\|_2\leq \gamma_m} |\|f_{j,k}\|^2_m-\|f_{j,k}\|^2_2|\geq \delta)\\
	&\leq \delta+m\phi(\ell)+P_0(\sup_{f_{j,k}\in B_{\mathcal H}(1), \|f_{j,k}\|_2\leq t} |\|f_{j,k}\|^2_m-\|f_{j,k}\|^2_2|\geq \delta).
	\end{align*}
	We choose $\delta$ to be $\frac{\gamma_m^2}{10}$, then using~\eqref{lemma6ind}, we have the upper bound
	\begin{equation*}
	\frac{\gamma_m^2}{10}+m\phi(\ell)+P_0(\sup_{f\in B_H(1), \|f\|_2\leq t} |\|f\|^2_m-\|f\|^2_2|\geq \frac{\gamma_m^2}{10})
	\leq \frac{\gamma_m^2}{10}+m\phi(\ell)+\tilde{c}_1\exp(-\tilde{c}_2m\gamma_m^2).
	\end{equation*}
	We require $m\gamma^2_m=\Omega(-\log(\gamma_m))$, which is the same as \cite{raskutti12}. Based on our assumptions, $m\phi(\ell)=m(m^{2/r_\phi})^{-r_\phi}=m^{-1}=o(\gamma_m^2)$ since $m\gamma_m^2\rightarrow \infty$ as $m\rightarrow \infty$. Therefore, for sufficiently large $m$, that expectation is bounded by $\frac{\gamma_m^2}{8}$. That completes the proof.
	
	For the follow-up statement, condition on event $\mathcal B_{m,T}$, for any $g_{j,k}\in B_{\mathcal H}(1)$ with $\|g_{j,k}\|_2\geq \gamma_m$, we have $h_{j,k}=\gamma_m\frac{g_{j,k}}{\|g_{j,k}\|_2}$ is in $B_{\mathcal H}(1)$ and $\|h_{j,k}\|_2\leq \gamma_m$. Therefore, we have 
	$$\left|\|\gamma_m\frac{g_{j,k}}{\|g_{j,k}\|_2}\|_T-\|\gamma_m\frac{g_{j,k}}{\|g_{j,k}\|_2}\|_2\right|\leq \frac{\gamma_m}{2},$$
	which implies
	$$|\|g_{j,k}\|_T-\|g_{j,k}\|_2|\leq \frac{1}{2}\|g_{j,k}\|_2.$$
\end{proof}
\begin{proof}[{\bf Proof of Lemma \ref{lemma:newmultibound}}]
	We follow the outline of proof for Lemma~\ref{lemma:multibound}. The only difference is here is the proof for showing
	\begin{equation*}
	\sup_{g_j\in 2\mathcal F_j, \|g_j\|_2\leq \delta_{m,j}}|\frac{1}{T}\sum_{t=1}^T\ell_\tau(g_j(X_t))-\mathbb E[\ell_\tau(g(X_t))]|\leq \frac{\delta_{m,j}^2}{4},
	\end{equation*}
	with high probability for $\phi$ mixing process with $0.781\leq r\leq 2$. 
	
	To show that, we use Lemma~\ref{McD} as in the proof of Lemma~\ref{lemma:newunivariate} and define
	 $$h(X)=\sup_{g_j\in 2\mathcal F_j, \|g_j\|_2\leq \delta_{m,j}}|\frac{1}{T}\sum_{t=1}^T\ell_\tau(g_j(X_t))-\mathbb E[\ell_\tau(g_j(X_1))]|.$$
	 We have
	 $$|h(X)-h(Y)|\leq \sup_{g_j\in 2\mathcal F_j, \|g_j\|_2\leq \delta_{m,j}}|\frac{1}{T}\sum_{t=1}^T(\ell_\tau(g_j(X_t))-\ell_\tau(g_j(Y_t)))|\leq \frac{\tau^2}{T}dist_{\mathcal HM}(X,Y),$$
	 which give us
	 $$\mathbb P(|h(X)-\mathbb Eh(X)|\geq \frac{\delta_{m,j}^2}{8})\leq O(exp(-\min(T^{2r_\phi-1},T)\frac{\delta_{m,j}^4}{\tau^4}))\leq c_2\exp(-c_3(m\delta_{m,j}^2)^2),$$
	 following the same analyses as in the proof of Lemma~\ref{lemma:newunivariate}. 
	 
	 As in the proof of Lemma~\ref{lemma:newunivariate}, we then need to show that for sufficient large $m$,
	  \begin{equation*}
	  E\sup_{g_j\in 2\mathcal F_j, \|g_j\|_2\leq \delta_{m,j}}|\frac{1}{T}\sum_{t=1}^T\ell_\tau(g_j(X_t))-\mathbb E[\ell_\tau(g_j(X))]|\leq \frac{\delta_{m,j}^2}{8}.
	  \end{equation*}
	  Using the same facts and results as we mentioned in the proof of Theorem~\ref{thm:univariate} and Lemma~\ref{lemma:newunivariate}, we show the upper bound in following.
	  \begin{align*}
	  &\mathbb E\sup_{g_j\in 2\mathcal F_j, \|g_j\|_2\leq \delta_{m,j}}|\frac{1}{T}\sum_{t=1}^T\ell_\tau(g_j(X_t))-\mathbb E[\ell_\tau(g_j(X_1))]|\\
	  &\leq \mathbb E\sup_{g_j\in 2\mathcal F_j, \|g_j\|_2\leq \delta_{m,j}}|\frac{1}{m}\sum_{a=1}^m\ell_\tau(g_j(X_{a,\ell}))-\mathbb E[\ell_\tau(g_j(X_{a,\ell}))]|\\
	  &\leq \delta+2\tau^2P(\sup_{g_j\in 2F_j, \|g_j\|_2\leq \delta_{m,j}}|\frac{1}{m}\sum_{a=1}^m\ell_\tau(g_j(X_{a,\ell}))-\mathbb E[\ell_\tau(g_j(X))]|\geq \delta)\\
	  &\leq \delta+2\tau^2m\phi(\ell)+2\tau^2\mathbb P_0(\sup_{g_j\in 2\mathcal F_j, \|g_j\|_2\leq \delta_{m,j}}|\frac{1}{m}\sum_{a=1}^m\ell_\tau(g_j(X_{a,\ell}))-\mathbb E[\ell_\tau(g_j(X))]|\geq \delta)\\
	  &\leq\frac{\delta_{m,j}^2}{12}+2\tau^2m\phi(\ell)+2\tau^2\mathbb P_0(\sup_{g_j\in 2\mathcal F_j, \|g_j\|_2\leq \delta_m}|\frac{1}{m}\sum_{a=1}^m\ell_\tau(g_j(X_{a,\ell}))-\mathbb E[\ell_\tau(g_j(X))]|\geq \frac{\delta_{m,j}^2}{12})\\
	  &\leq \frac{\delta_{m,j}^2}{12}+2\tau^2m\phi(\ell)+2\tau^2\tilde{c}_1\exp{-(\tilde{c}_2m\delta_{m,j}^2)},
	  \end{align*}
	  which is bounded by $\frac{\delta_{m,j}^2}{8}$ for sufficiently large $m$, using similar arguments as in the proof for Lemma \ref{lemma:unibound}. That completes the proof.
\end{proof}
\end{document}